\documentclass[10pt,twocolumn,letterpaper]{article}

\usepackage{iccv}
\usepackage{times}
\usepackage{epsfig}
\usepackage{graphicx}
\usepackage{amsmath}
\usepackage{amssymb}
\usepackage{xfrac}
\usepackage[mathscr]{euscript}
\usepackage{appendix}
\usepackage{algorithm}
\usepackage{algpseudocode}
\usepackage{cancel}

\usepackage{color,soul}
\usepackage{tabularx}
\usepackage{subcaption}
\usepackage{mathtools}
\usepackage{paralist} 
\usepackage{colortbl}
\usepackage{multirow}
\usepackage[table]{xcolor}

\usepackage[accsupp]{axessibility}

\usepackage[breaklinks=true,bookmarks=false]{hyperref}

\iccvfinalcopy 

\def\iccvPaperID{****} 

\ificcvfinal\fi

\newcommand{\ourtitle}{Zip-NeRF: Anti-Aliased Grid-Based Neural Radiance Fields}

\newcommand{\powerladder}{\mathcal{P}}
\newcommand{\power}{\lambda}

\definecolor{yellow}{rgb}{1, 1, 0.7}
\definecolor{orange}{rgb}{1, 0.85, 0.7}
\definecolor{red}{rgb}{1, 0.7, 0.7}

\definecolor{lightyellow}{rgb}{1,1, 0.8}

\definecolor{wincolor}{rgb}{0.85, 0.0, 0.0}

\definecolor{darkyellow}{rgb}{0.8, 0.8, 0.5}
\definecolor{darkred}{rgb}{0.7, 0.3, 0.3}
\definecolor{darkgreen}{rgb}{0.3, 0.7, 0.3}
\definecolor{blue}{rgb}{0, 0, 1.0}
\definecolor{green}{rgb}{0, 1.0, 0}
\definecolor{pink}{rgb}{1, 0.4, 0.7}

\let\originalleft\left
\let\originalright\right
\renewcommand{\left}{\mathopen{}\mathclose\bgroup\originalleft}
\renewcommand{\right}{\aftergroup\egroup\originalright}

\newcommand{\norm}[1]{\left\lVert#1\right\rVert}

\newcommand{\sdistance}{s}

\newcommand{\baseradius}{\dot r}

\newcommand{\zval}{t}

\newcommand{\proposal}[1]{\hat{#1}}

\newcommand{\myparagraph}[1]{\paragraph{#1}}

\newcommand{\zc}{\zval_\mu}
\newcommand{\zd}{\zval_\delta}

\newcommand{\proploss}{\mathcal{L}_{\mathrm{prop}}}

\newcommand{\downweight}{\omega}
\newcommand{\isostd}{\sigma}
\newcommand{\grid}{V}
\newcommand{\gridsize}{n}
\newcommand{\gridchannels}{c}
\newcommand{\contract}{\mathcal{C}}
\newcommand{\jacobian}{\mathbf{J}}
\newcommand{\initval}{\grid_{\mathit{init}}}
\newcommand{\stopgrad}{\cancel{\nabla}}

\newcommand{\level}{\ell}

\renewcommand{\t}{\mathbf{x}}
\newcommand{\y}{\mathbf{y}}
\newcommand{\tp}{\t_r}
\newcommand{\yp}{\y_r}
\newcommand{\dy}{\y'}
\newcommand{\dyp}{\y''}
\newcommand{\idx}{\mathit{sortidx}}

\begin{document}

\title{\ourtitle}

\author{
Jonathan T. Barron
\quad\quad
Ben Mildenhall
\quad\quad
Dor Verbin
\quad\quad
Pratul P. Srinivasan
\quad\quad
Peter Hedman
\\
\vspace{4mm}
{Google Research}
}

\ificcvfinal\thispagestyle{empty}\fi

\newcommand{\mywidth}{0.192\linewidth}
\newcommand{\toptrim}{0in}

\twocolumn[{
\renewcommand\twocolumn[1][]{#1}
\maketitle
\begin{center}
    \begin{tabular}{@{}c@{\,\,}c@{\,\,}c@{\,\,}c@{\,\,}c@{}}
        \includegraphics[trim=0 0in 0 \toptrim, clip, width=\mywidth]{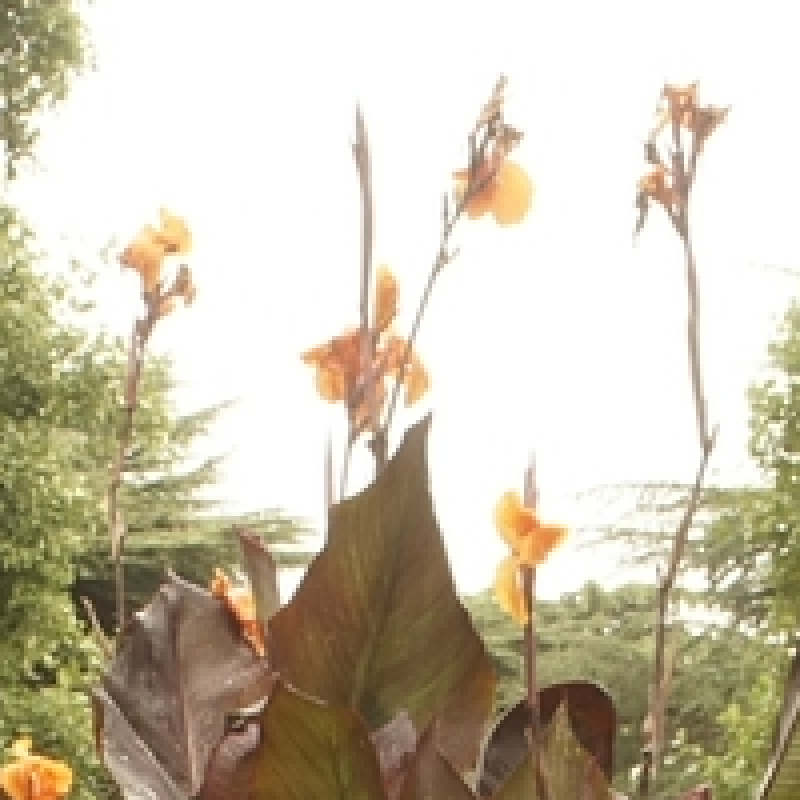} &
        \includegraphics[trim=0 0in 0 \toptrim, clip, width=\mywidth]{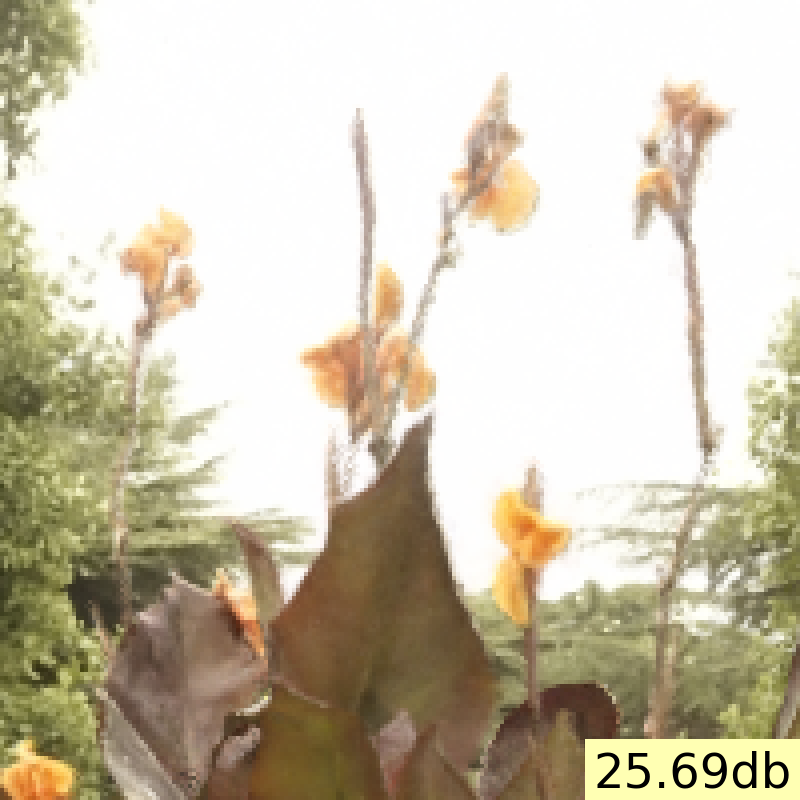} &
        \includegraphics[trim=0 0in 0 \toptrim, clip, width=\mywidth]{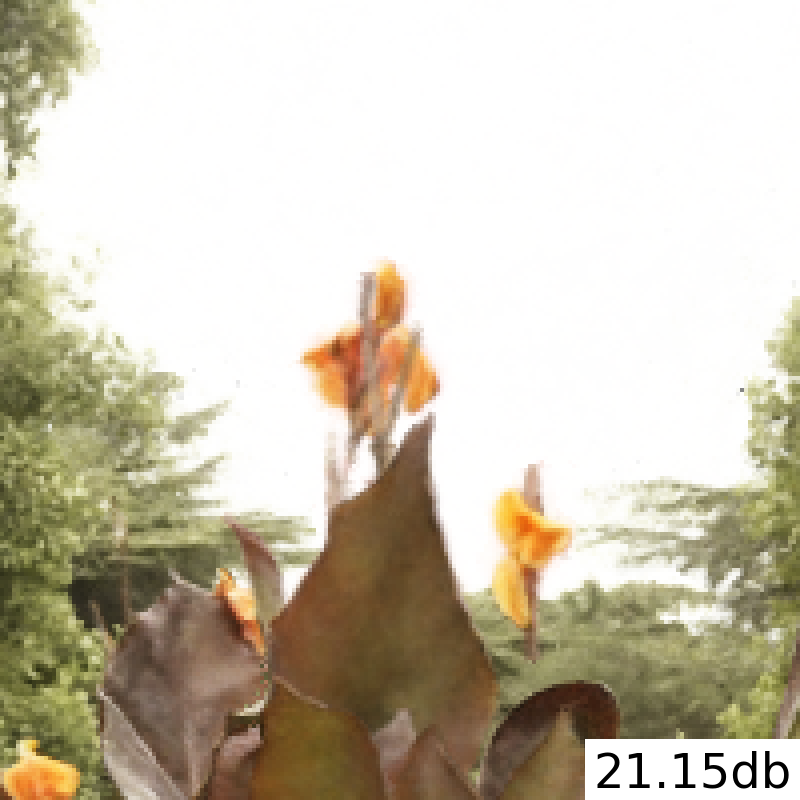} &
        \includegraphics[trim=0 0in 0 \toptrim, clip, width=\mywidth]{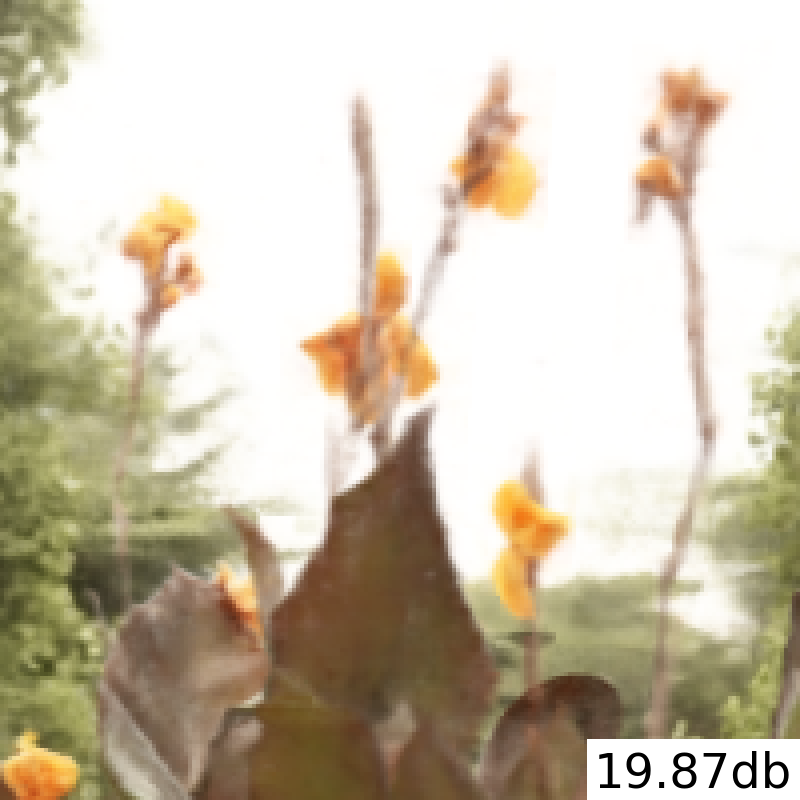} &
        \includegraphics[trim=0 0in 0 \toptrim, clip, width=\mywidth]{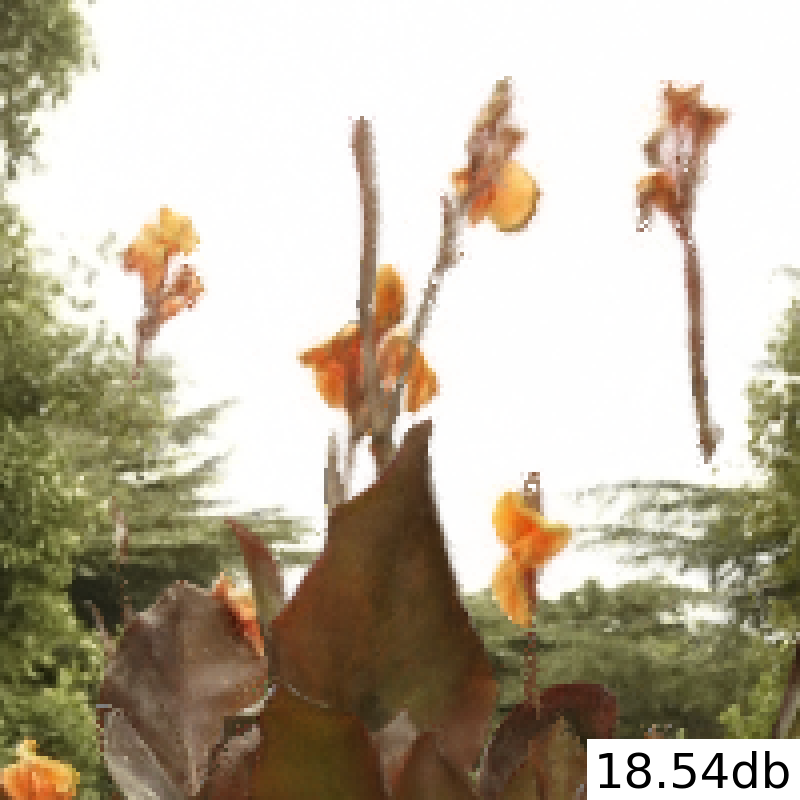} \\
        \includegraphics[trim=0 0in 0 \toptrim, clip, width=\mywidth]{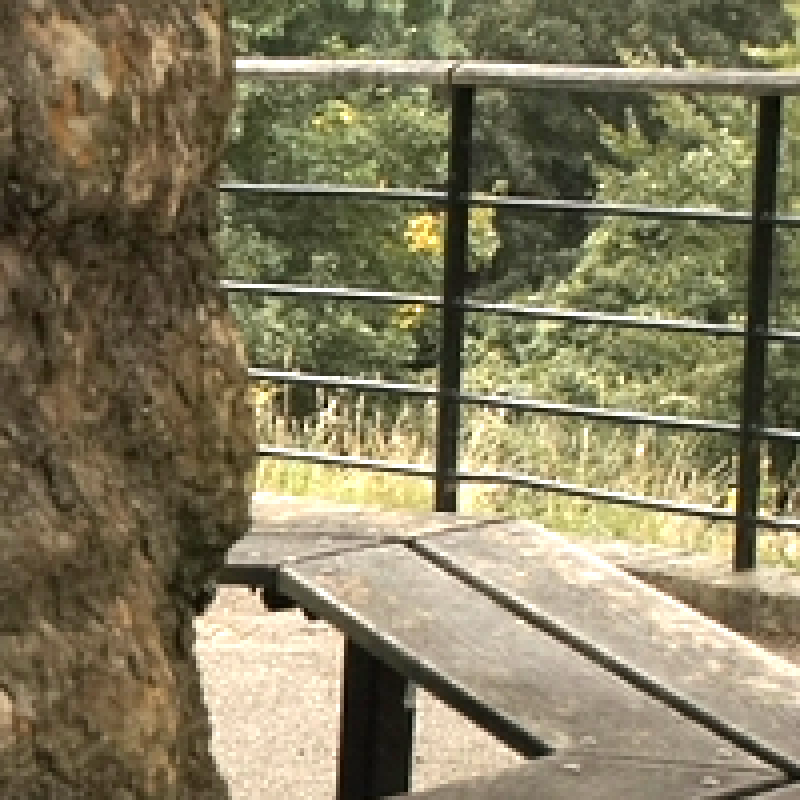} &
        \includegraphics[trim=0 0in 0 \toptrim, clip, width=\mywidth]{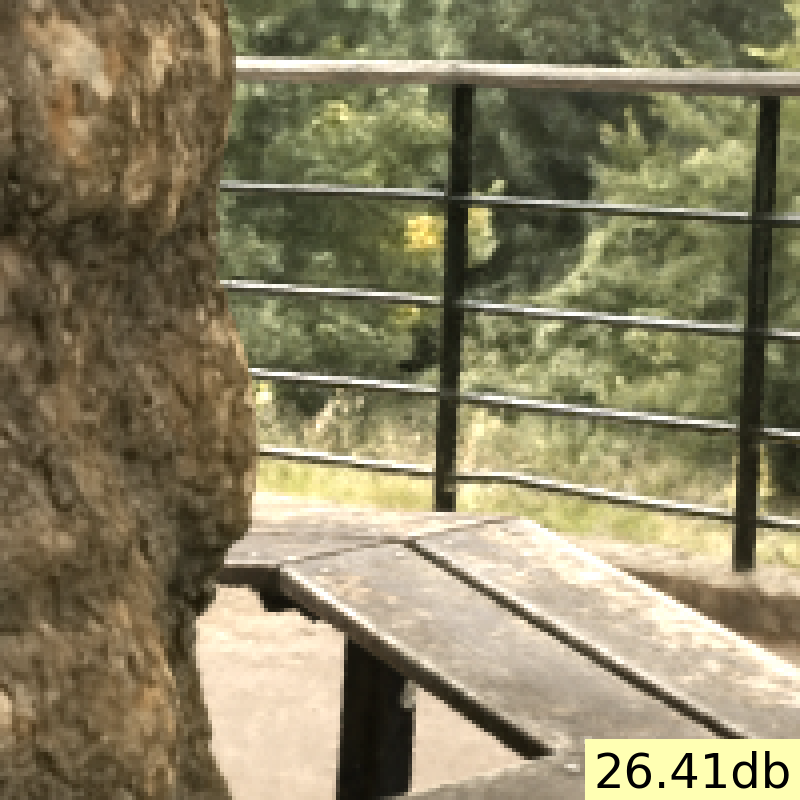} &
        \includegraphics[trim=0 0in 0 \toptrim, clip, width=\mywidth]{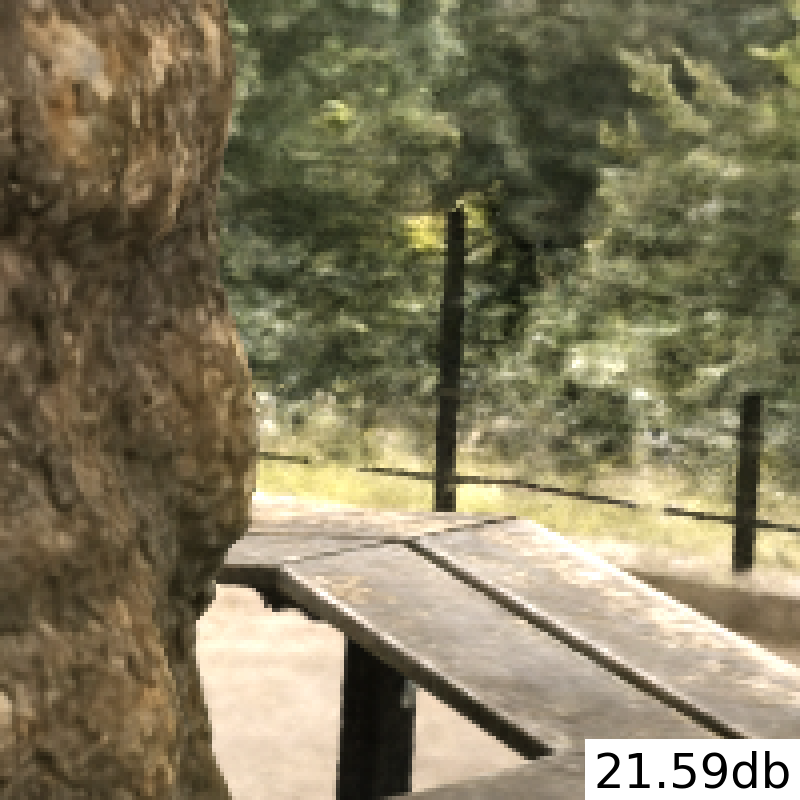} &
        \includegraphics[trim=0 0in 0 \toptrim, clip, width=\mywidth]{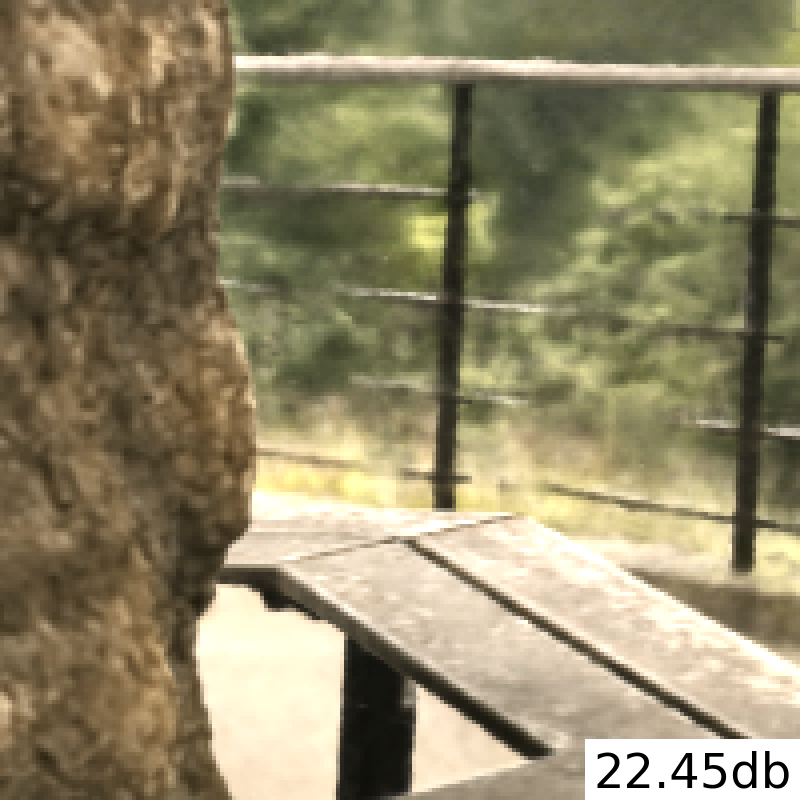} &
        \includegraphics[trim=0 0in 0 \toptrim, clip, width=\mywidth]{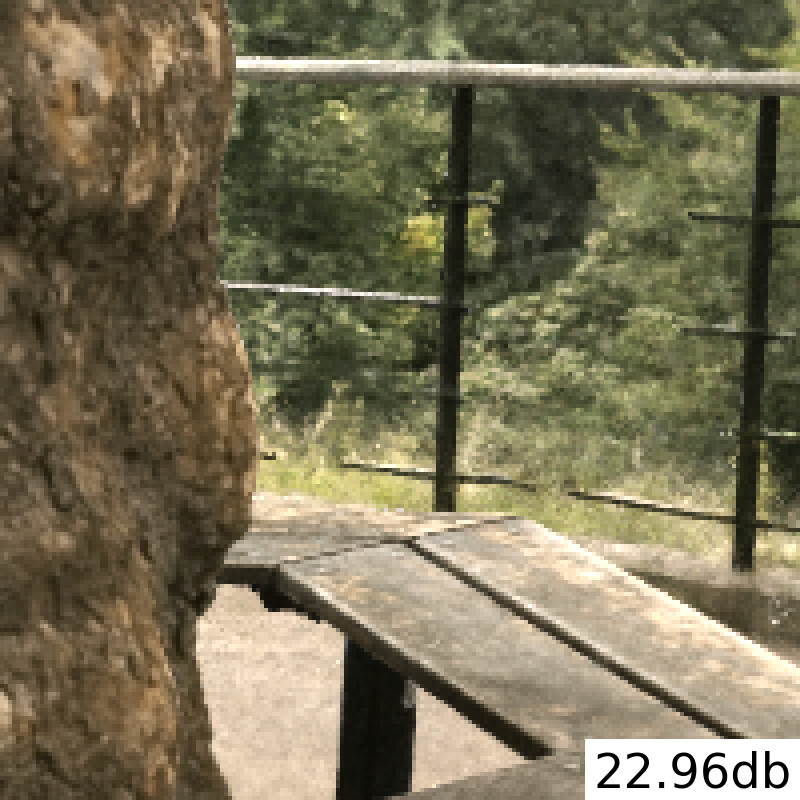} \\[-0.75ex]
        \scriptsize (a) Ground Truth  & \scriptsize (b) Our Model & \scriptsize (c) mip-NeRF 360~\cite{barron2022mipnerf360} & \scriptsize (d) iNGP~\cite{muller2022instant} & \scriptsize (e) mip-NeRF 360 + iNGP 
    \end{tabular}
\captionof{figure}{
(a) Test-set images from the mip-NeRF 360 dataset~\cite{barron2022mipnerf360} with renderings from (b) our model and (c, d, e) three state-of-the-art baselines.
Our model accurately recovers thin structures and finely detailed foliage, while the baselines either oversmooth or exhibit aliasing in the form of jaggies and missing scene content.
PSNR values for each patch are inset.
}\label{fig:single_results}
\end{center}
}]

\begin{abstract}
Neural Radiance Field training can be accelerated through the use of grid-based representations in NeRF's learned mapping from spatial coordinates to colors and volumetric density.
However, these grid-based approaches lack an explicit understanding of scale and therefore often introduce aliasing, usually in the form of jaggies or missing scene content.
Anti-aliasing has previously been addressed by mip-NeRF 360, which reasons about sub-volumes along a cone rather than points along a ray, but this approach is not natively compatible with current grid-based techniques.
We show how ideas from rendering and signal processing can be used to construct a technique that combines mip-NeRF 360 and grid-based models such as Instant NGP to yield error rates that are 8\% -- 77\% lower than either prior technique, and that trains $24\times$ faster than mip-NeRF 360.
\end{abstract}

In Neural Radiance Fields (NeRF), a neural network is trained to model a volumetric representation of a 3D scene such that novel views of that scene can be rendered via ray-tracing~\cite{mildenhall2020}. NeRF has proven to be an effective tool for tasks such as view synthesis~\cite{martinbrualla2020nerfw}, generative media~\cite{poole2022dreamfusion}, robotics~\cite{yen2022nerfsupervision}, and computational photography~\cite{mildenhall2022rawnerf}.

The original NeRF model used a multilayer perceptron (MLP) to parameterize the mapping from spatial coordinates to colors and densities. Though compact and expressive, MLPs are slow to train, and recent work has accelerated training by replacing or augmenting MLPs with voxel-grid-like datastructures~\cite{chen2022tensorf, karnewar2022relu, sun2022direct, yu2021plenoxels}. One example is Instant NGP (iNGP), which uses a pyramid of coarse and fine grids (the finest of which are stored using a hash map) to construct learned features that are processed by a tiny MLP, enabling greatly accelerated training~\cite{muller2022instant}.

\begin{figure*}[t!]
    \centering
    \begin{tabular}{@{}c@{\,}c@{\,}|@{\,}c@{\,\,}c@{\,\,}c@{\,\,}c@{}}
    \rotatebox{90}{\scriptsize \quad\,\, Fine $\longleftrightarrow$ Coarse \hspace{0.05in} Method} &
        \includegraphics[trim=0 0.1in 0 0.1in, clip, width=0.19\linewidth]{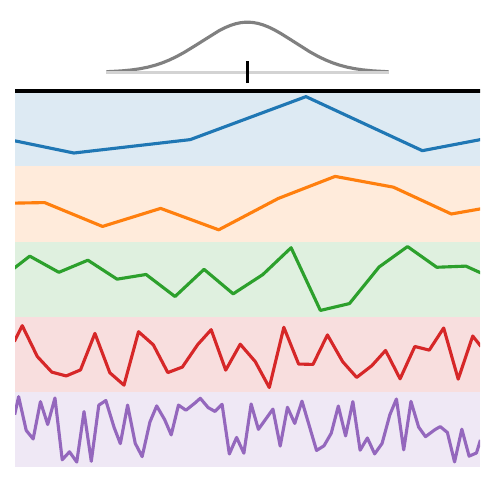} &
        \includegraphics[trim=0 0.1in 0 0.1in, clip, width=0.19\linewidth]{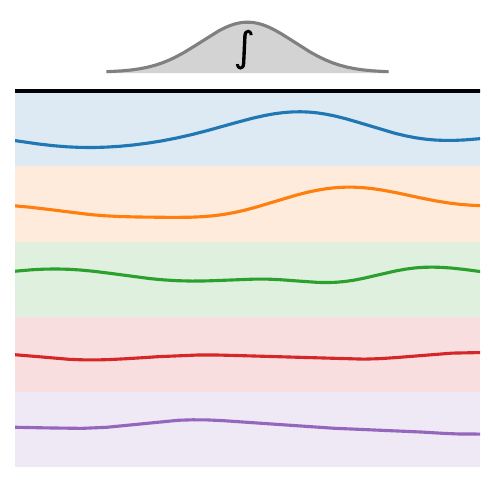} &
        \includegraphics[trim=0 0.1in 0 0.1in, clip, width=0.19\linewidth]{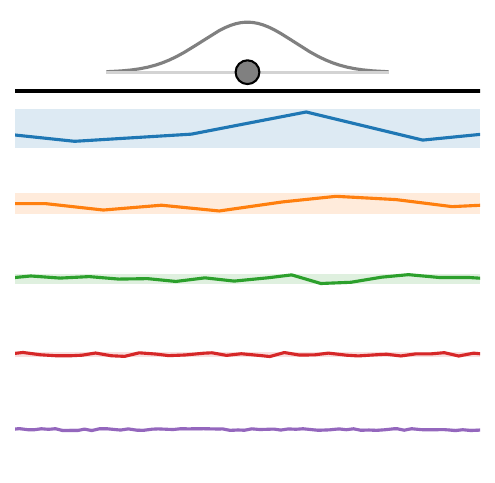} &
        \includegraphics[trim=0 0.1in 0 0.1in, clip, width=0.19\linewidth]{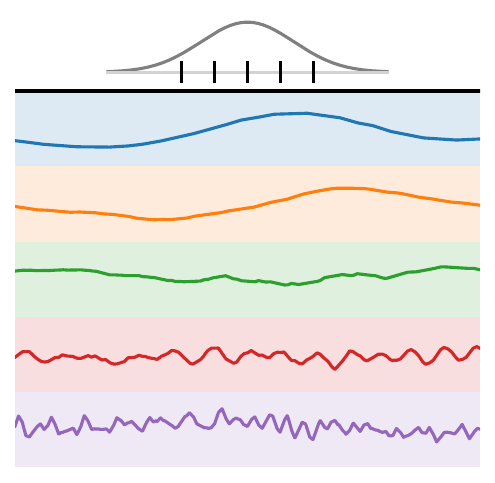} &
        \includegraphics[trim=0 0.1in 0 0.1in, clip, width=0.19\linewidth]{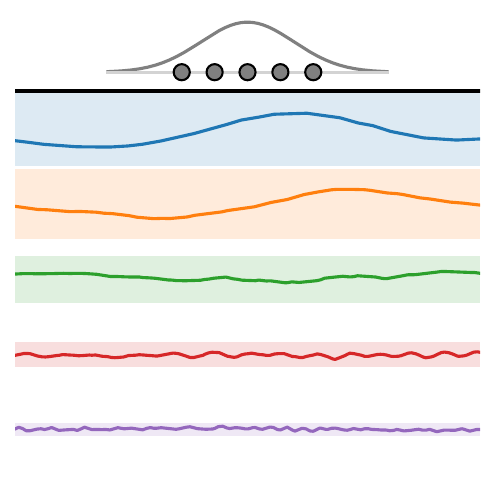} \\[-0.75ex]
        & \scriptsize (a) Input / Baseline, 14.6 dB & \scriptsize (b) Ground-truth & \scriptsize (c) Downweighting, 24.3 dB & \scriptsize (d) Multisampling, 25.2 dB & \scriptsize (e) Our Approach, \hl{26.5 dB}
    \end{tabular}
    \caption{
    Here we show a toy 1-dimensional iNGP~\cite{muller2022instant} with 1 feature per scale.
    Each subplot represents a different strategy for querying the iNGP at all coordinates along the $x$ axis ---
    imagine a Gaussian moving left to right, where each line is the iNGP feature for each coordinate, and where each color is a different scale in the iNGP.
    (a) The naive solution of querying the Gaussian's mean results in features with piecewise-linear kinks, where the high frequencies past the bandwidth of the Gaussian are large and inaccurate.
    (b) The true solution, obtained by convolving the iNGP features with a Gaussian --- an intractable solution in practice --- results in coarse features that are smooth but informative and fine features that are near 0.
    (c) We can suppress unreliable high frequencies by downweighing them based on the scale of the Gaussian (color bands behind each feature indicate the downweighting), but this results in unnaturally sharp discontinuities in coarse features. 
    (d) Alternatively, supersampling produces reasonable coarse scales features but erratic fine-scale features.
    (e) We therefore multisample isotropic sub-Gaussians (5 shown here) and use each sub-Gaussian's scale to downweight frequencies.
    }
    \label{fig:control}
\end{figure*}

In addition to being slow, the original NeRF model was also aliased: NeRF reasons about individual points along a ray, which results in ``jaggies'' in rendered images and limits NeRFs ability to reason about scale. Mip-NeRF~\cite{barron2021mipnerf} resolved this issue by casting cones instead of rays, and by featurizing the entire volume within a conical frustum for use as input to the MLP. Mip-NeRF and its successor mip-NeRF 360~\cite{barron2022mipnerf360} showed that this approach enables highly accurate rendering on challenging real-world scenes.

Regrettably, the progress made on these two problems of fast training and anti-aliasing are, at first glance, incompatible with each other. This is because mip-NeRF's anti-aliasing strategy depends critically on the use of positional encoding~\cite{tancik2020fourfeat, vaswani2017attention} to featurize a conical frustum into a discrete feature vector, but current grid-based approaches do not use positional encoding, and instead use learned features that are obtained by interpolating into a hierarchy of grids at a single 3D coordinate.
Though anti-aliasing is a well-studied problem in rendering~\cite{cook1986stochastic, crow1984summed, shirley1991discrepancy, williams83}, most approaches do not generalize naturally to inverse-rendering in grid-based NeRF models like iNGP.

In this work, we leverage ideas from multisampling, statistics, and signal processing to integrate iNGP's pyramid of grids into mip-NeRF 360's framework.
We call our model ``Zip-NeRF'' due to its speed, its similarity with mip-NeRF, and its ability to fix zipper-like aliasing artifacts. On the mip-NeRF 360 benchmark~\cite{barron2022mipnerf360}, 
Zip-NeRF reduces error rates by as much as 19\% and trains $24 \times$ faster than the previous state-of-the-art. On our multiscale variant of that benchmark, which more thoroughly measures aliasing and scale, Zip-NeRF reduces error rates by as much as 77\%.

\section{Preliminaries}

Mip-NeRF 360 and Instant NGP (iNGP) are both NeRF-like~\cite{mildenhall2020}: A pixel is rendered by casting a 3D ray and featurizing locations at distances $t$ along the ray, and those features are fed to a neural network whose outputs are alpha-composited to render a color. Training consists of repeatedly casting rays corresponding to pixels in training images and minimizing (via gradient descent) the difference between each pixel's rendered and observed colors.

Mip-NeRF 360 and iNGP differ significantly in how coordinates along a ray are parameterized. In mip-NeRF 360, a ray is subdivided into a set of intervals $[t_i, t_{i+1})$, each of which represents a conical frustum whose shape is approximated with a multivariate Gaussian, and the expected positional encoding with respect to that Gaussian is used as input to a large MLP~\cite{barron2021mipnerf}.
In contrast, iNGP trilinearly interpolates into a hierarchy of differently-sized 3D grids to produce feature vectors for a small MLP~\cite{muller2022instant}.
Our model combines mip-NeRF 360's overall framework with iNGP's featurization approach, but naively combining these two approaches introduces two forms of aliasing:
\begin{enumerate}
\item  Instant NGP's feature grid approach is incompatible with mip-NeRF 360's scale-aware integrated positional encoding technique, so the features produced by iNGP are aliased with respect to spatial coordinates and thus produce aliased renderings. In Section~\ref{sec:spatial_aliasing}, we address this by introducing a multisampling-like solution for computing prefiltered iNGP features.
\item Using iNGP dramatically accelerates training, but this reveals a problem with mip-NeRF 360's online distillation approach that causes highly visible ``$z$-aliasing'' (aliasing along a ray), wherein scene content erratically disappears as the camera moves. In Section~\ref{sec:z_aliasing} we resolve this with a new loss function that prefilters along each ray when computing the loss function used to supervise online distillation.
\end{enumerate}

\section{Spatial Anti-Aliasing}
\label{sec:spatial_aliasing}

Mip-NeRF uses features that approximate the integral over the positional encoding of the coordinates within a sub-volume, which is a conical frustum.
This results in Fourier features whose amplitudes are small when the feature sinusoid's period is larger than the standard deviation of the Gaussian --- the features express the spatial location of the sub-volume \emph{only} at the wavelengths that are larger than the size of the sub-volume.
Because this feature encodes both position and scale, the MLP that consumes it is able to learn a multi-scale representation of the 3D scene that renders anti-aliased images.
Grid-based representations like iNGP do not natively allow for sub-volumes to be queried, and instead use trilinear interpolation at a single point to construct features for use in an MLP, which results in learned models that cannot reason about scale or aliasing.

\begin{figure}[t]
    \centering
    \begin{tabular}{@{}c@{\,\,}c@{}}
        \includegraphics[width=0.49\linewidth]{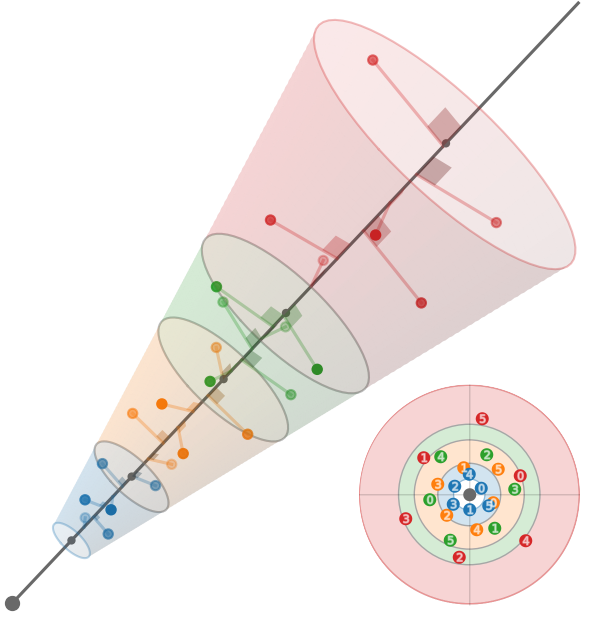} &
        \includegraphics[width=0.49\linewidth]{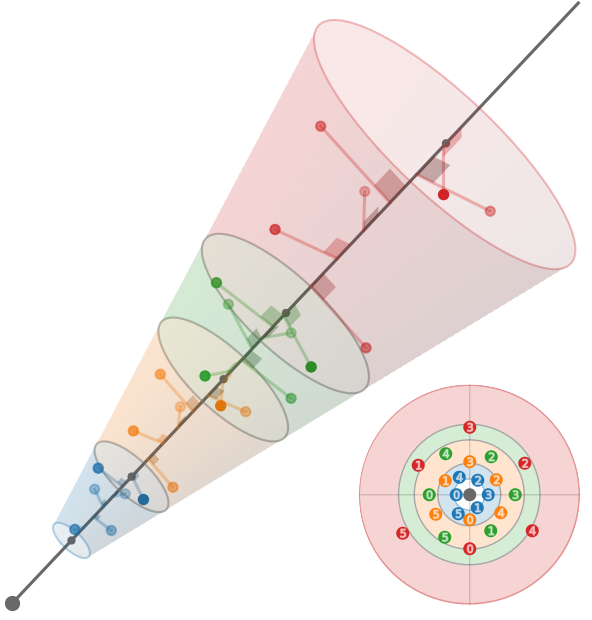} \\[-0.75ex]
        \small (a) Training & \small (b) Testing
    \end{tabular}
    \vspace{-0.1in}
    \caption{
    Here we show a toy 3D ray with an exaggerated pixel width (viewed along the ray as an inset) divided into 4 frustums denoted by color. 
    We multisample each frustum with a hexagonal pattern that matches the frustum's first and second moments. Each pattern is rotated around the ray and flipped along the ray (a) randomly when training and (b) deterministically when rendering.
    \label{fig:multisamples}
    }
\end{figure}

We resolve this by turning each conical frustum into a set of isotropic Gaussians, using a combination of multisampling and feature-downweighting: The anisotropic sub-volume is first converted into a set of points that approximate its shape, and then each point is assumed to be an isotropic Gaussian with some scale. This isotropic assumption lets us approximate the true integral of the feature grid over a sub-volume by leveraging the fact that the values in the grid are zero-mean. By averaging these downweighted features, we obtain scale-aware prefiltered features from an iNGP grid. See Figure~\ref{fig:control} for a visualization.

Anti-aliasing is well explored in the graphics literature.
Mip mapping~\cite{williams83} (mip-NeRF's namesake) precomputes a datastructure that enables fast anti-aliasing, but it is unclear how this approach can be applied to iNGP's hash-based datastructures.
Supersampling techniques~\cite{cook1986stochastic} adopt a brute-force approach to anti-aliasing and use a large number of samples; we will demonstrate that this is less effective and more expensive than our approach.
Multisampling techniques~\cite{greene1986creating} construct a small set of samples, and then pool information from those multisamples into an aggregate representation that is provided to an expensive rendering process ---  a strategy that resembles our approach.

\myparagraph{Multisampling} Following mip-NeRF~\cite{barron2021mipnerf}, we assume each pixel corresponds to a cone with radius $\baseradius t$, where $t$ is distance along the ray. Given an interval along the ray $[t_0, t_1)$, we would like to construct a set of multisamples that approximate the shape of that conical frustum.
We use a 6-point hexagonal pattern whose angles $\theta_j$ are:
\begin{equation}
  \boldsymbol{\theta} = \left[\, 0, \sfrac{2\pi}{3}, \sfrac{4\pi}{3}, \sfrac{3\pi}{3}, \sfrac{5\pi}{3}, \sfrac{\pi}{3} \,\right]\,,
\end{equation}
which are linearly-spaced angles around one rotation of a circle, permuted to give a pair of triangles that are shifted by 60 degrees. The distances along the ray $t_j$ are:
\begin{gather}
t_j = t_0 + \frac{\zd \left(\! t_1^2 + 2\zc^2 + \frac{3}{\sqrt{7}}\left( \frac{2j}{5}-1 \right) \sqrt{\!\left(\zd^2 - \zc^2 \right)^2\! + 4 \zc^4}\right)}{\zd^2 + 3 \zc^2} \nonumber \\
\text{where} \,\, \zc = (\zval_0 + \zval_1) / 2\,,\,\,\, \zd = (\zval_1 - \zval_0) / 2
\end{gather}
which are linearly-spaced values in $[t_0, t_1)$, shifted and scaled to concentrate mass near the far base of the frustum.
Our multisample coordinates relative to the ray are:
\begin{gather}
\left\{ {\begin{bmatrix} \baseradius t_j \cos(\theta_j)/\sqrt{2} \\ \baseradius t_j \sin(\theta_j)/\sqrt{2} \\ t_j \end{bmatrix} \,\Bigg|\, \begin{matrix} j = 0, 1, \ldots, 5 \end{matrix} } \right\}\,.
\end{gather}
These 3D coordinates are rotated into world coordinates by multiplying them by an orthonormal basis whose third vector is the ray direction (and whose first two vectors are an arbitrary frame that is perpendicular to the ray) and then shifted by the ray origin.
By construction, the sample means and variances (along the ray and perpendicular to the ray) of these multisamples exactly match those of the conical frustum, analogously to mip-NeRF Gaussians~\cite{barron2021mipnerf} (see the supplement for details). During training, we randomly rotate and flip each pattern, and during rendering we deterministically flip and rotate every other pattern by 30 degrees.
See Figure~\ref{fig:multisamples} for visualizations of both strategies.

We use these 6 multisamples $\{ \mathbf{x}_j \}$ as the means of isotropic Gaussians each with a standard deviation of $\sigma_j$. We set $\sigma_j$ to $\baseradius t_j / \sqrt{2}$ scaled by a hyperparameter, which is $0.5$ in all experiments. Because iNGP grids require input coordinates to lie in a bounded domain, we apply the contraction function from mip-NeRF 360~\cite{barron2022mipnerf360}.
Since these Gaussians are isotropic, we can compute the scale factor induced by this contraction using an efficient alternative to the Kalman filter approach used by mip-NeRF 360; see the supplement for details.

\begin{figure*}[h!]
    \centering
    \includegraphics[trim=0 7.6in 0 0in, clip, width=\linewidth]{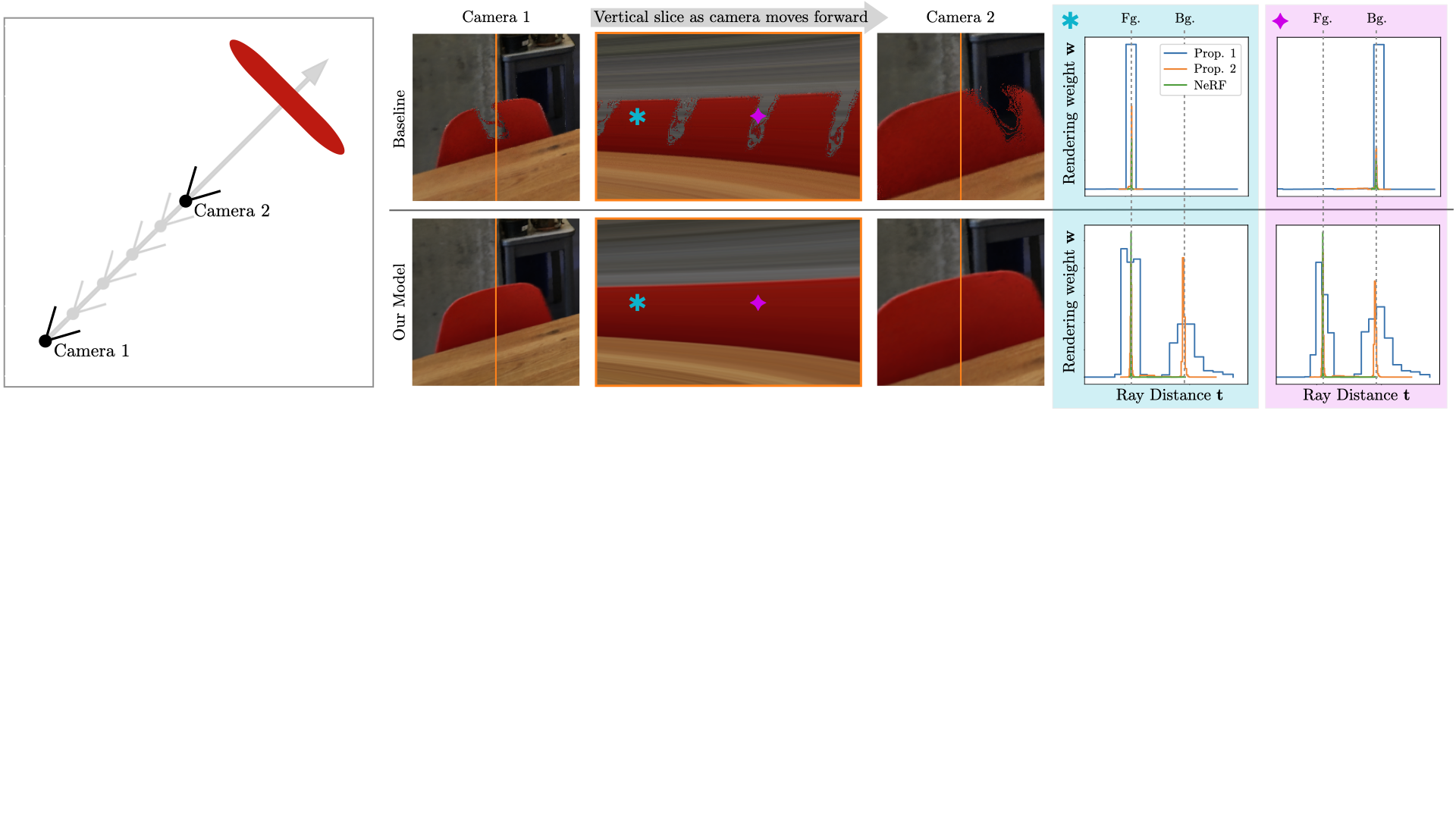}
    \vspace{-0.25in}
    \caption{
    Here we visualize the problem of $z$-aliasing. Left: we have a scene where 2 training cameras face a narrow red chair in front of a gray wall.
    Middle: As we sweep a test camera between those training cameras,  we see that the baseline algorithm (top) ``misses'' or ``hits'' the chair depending on its distance and therefore introduces tearing artifacts, while our model (bottom) consistently ``hits'' the chair to produce artifact-free renderings.
    Left: This is because the baseline (top) has learned non-smooth proposal distributions due to aliasing in its supervision, while our model (bottom) correctly predicts proposal distributions that capture both the foreground and the background at all depths due to our anti-aliased loss function.
    }
    \label{fig:zaliasing}
\end{figure*}

\myparagraph{Downweighting} Though multisampling is an effective tool for reducing aliasing, using naive trilinear interpolation for each multisample still results in aliasing at high frequencies, as can be seen in Figure~\ref{fig:control}(d). To anti-alias interpolation \emph{for each individual multisample}, we re-weight the features at each scale in a way that is inversely-proportional to how much each multisample's isotropic Gaussian fits within each grid cell: if the Gaussian is much larger than the cell being interpolated into, the interpolated feature is likely unreliable and should be downweighted. Mip-NeRF's IPE features have a similar interpretation.

In iNGP, interpolation into each grid or hash $\{ \grid_\level \}$ at coordinate $\mathbf{x}$ is done by scaling $\mathbf{x}$ by the grid's linear size $\gridsize_\level$ and performing trilinear interpolation into $\grid_\level$ to get a $\gridchannels_\ell$-length vector.
We instead interpolate a set of multisampled isotropic Gaussians with means $\{ \mathbf{x}_j \}$ and standard deviations $\{ \sigma_j \}$. By reasoning about Gaussian CDFs we can compute the fraction of each Gaussian's PDF that is inside the $[-\sfrac{1}{2\gridsize}, \sfrac{1}{2\gridsize}]^3$ cube in $\grid_\level$ being interpolated into as
a scale-dependent downweighting factor 
$\downweight_{j, \ell} = \operatorname{erf}(  1/\scriptstyle\sqrt{8 \isostd_j^2 \gridsize_\ell^2} \,\textstyle)$, where $\downweight_{j, \ell} \in [0, 1]$.
As detailed in Section~\ref{sec:details}, we impose weight decay on $\{ \grid_\ell \}$, which encourages the values in $\grid_\ell$ to be normally distributed and zero-mean.
This zero-mean assumption lets us approximate the expected grid feature with respect to each multisample's Gaussian as $\downweight_j \cdot \mathbf{f}_{j, \ell} + (1 - \downweight_j) \cdot \mathbf{0} = \downweight_j \cdot \mathbf{f}_{j, \ell}$.
With this, we can approximate the expected feature corresponding to the conical frustum being featurized by taking a weighted mean of each multisample's interpolated features:
\begin{equation}
\mathbf{f}_\level = \operatorname{mean}_j\left(\downweight_{j, \ell} \cdot \operatorname{trilerp}(\gridsize_\level \cdot \mathbf{x}_j ; \grid_\ell) \right)\,.
\end{equation}
This set of features $\{ \mathbf{f}_\ell \}$ is concatenated and provided as input to an MLP, as in iNGP.
We also concatenate featurized versions of $\{ \downweight_{j, \ell}\}$, see the supplement for details.

\section{Z-Aliasing and Proposal Supervision}
\label{sec:z_aliasing}

\begin{figure*}[t]
    \centering
    \includegraphics[trim=0 9.91in 1.3in 0, clip, width=0.98\linewidth]{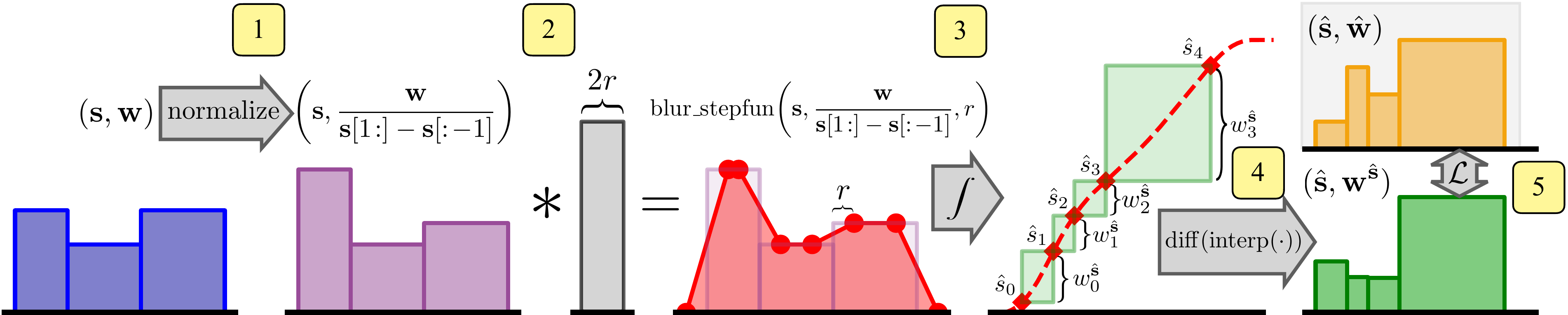}
    \caption{
    Computing our anti-aliased loss requires that we smooth and resample 
    a NeRF histogram $(\mathbf{s}, \mathbf{w})$ into the same set of endpoints as a proposal histogram $(\hat{\mathbf{s}}, \hat{\mathbf{w}})$, which we outline here. (1) We divide $\mathbf{w}$ by the size of each interval in $\mathbf{s}$ to yield a piecewise constant PDF that integrates to $\leq 1$. (2) We convolve that PDF with a rectangular pulse to obtain a piecewise linear PDF. (3) This PDF is integrated to produce a piecewise-quadratic CDF that is queried via piecewise quadratic interpolation at each location in $\hat{\mathbf{s}}$. (4) By taking the difference between adjacent interpolated values we obtain $\mathbf{w}^{\hat{\mathbf{s}}}$, which are the NeRF histogram weights $\mathbf{w}$ resampled into the endpoints of the proposal histogram $\hat{\mathbf{s}}$. (5) After resampling, we evaluate our loss $\proploss$ as an element-wise function of $\mathbf{w}^{\hat{\mathbf{s}}}$ and $\hat{\mathbf{w}}$, as they share a common coordinate space.
    \label{fig:zaliasing_solution}
    }
\end{figure*}

Though the previously-detailed multisampling and downweighting approach is an effective way to reduce spatial aliasing, there is an additional source of aliasing \emph{along the ray} that we must consider, which we will refer to as $z$-aliasing. This $z$-aliasing is due to mip-NeRF 360's use of a proposal MLP that learns to produce upper bounds on scene geometry: during training and rendering, intervals along a ray are repeatedly evaluated by this proposal MLP to generate histograms that are resampled by the next round of sampling, and only the final set of samples is rendered by the NeRF MLP. Mip-NeRF 360 showed that this approach significantly improves speed and rendering quality compared to prior strategies of learning one~\cite{barron2021mipnerf} or multiple~\cite{mildenhall2020} NeRFs that are all supervised using image reconstruction.
We observe that the proposal MLP in mip-NeRF 360 tends to learn a non-smooth mapping from input coordinates to output volumetric densities. This results in artifacts in which a ray ``skips'' scene content, which can be seen in Figure~\ref{fig:zaliasing}. Though this artifact is subtle in mip-NeRF 360, if we use an iNGP backend for our proposal network instead of an MLP (thereby increasing our model's ability to optimize rapidly) it becomes common and visually salient, especially when the camera translates along its $z$-axis.

\begin{figure}[b]
    \centering
    \begin{tabular}{@{}c@{\,\,}c@{}}
        \includegraphics[trim=0 0.125in 0 0, clip, width=0.49\linewidth]{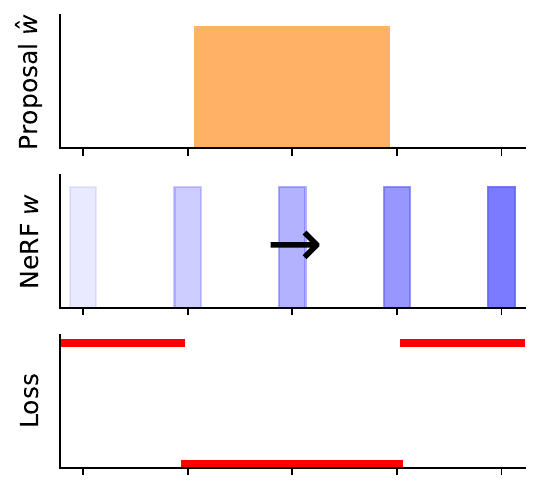} &
        \includegraphics[trim=0 0.125in 0 0, clip, width=0.457\linewidth]{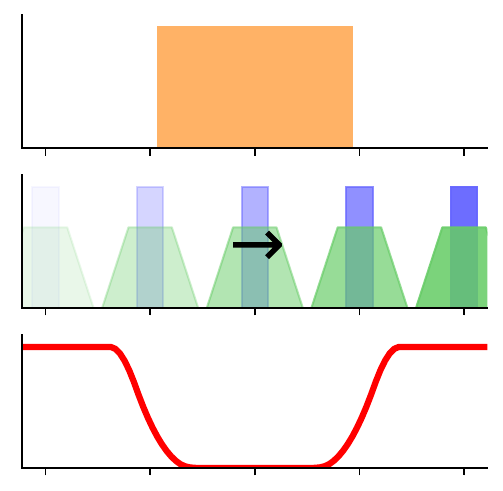} \\
        \scriptsize (a) mip-NeRF 360 & \scriptsize (b) Our Approach
    \end{tabular}
    \caption{
    Here we visualize proposal supervision for a toy setting where a narrow NeRF histogram (blue) translates along a ray relative to a coarse proposal histogram (orange). (a) The loss used by mip-NeRF 360 is piecewise constant, but (b) our loss is smooth because we blur NeRF histograms into piecewise linear splines (green). Our prefiltered loss lets us learn anti-aliased proposal distributions.
    \label{fig:zaliasing_losses}
    }
\end{figure}

The root cause of $z$-aliasing in mip-NeRF 360 is the ``interlevel loss'' used to supervise the proposal network, which assigns an equivalent loss to NeRF and proposal histogram bins whether their overlap is partial or complete --- proposal histogram bins are only penalized when they completely fail to overlap. To resolve this issue, we present an alternative loss that, unlike mip-NeRF 360's interlevel loss, is continuous and smooth with respect to distance along the ray. See Figure~\ref{fig:zaliasing_losses} for a comparison of both loss functions.

\myparagraph{Blurring a Step Function}
To design a loss function that is smooth with respect to distance along the ray, we must first construct a technique for turning a piecewise constant step function into a continuous piecewise-linear function. Smoothing a discrete 1D signal is trivial, and requires just discrete convolution with a box filter (or a Gaussian filter, \etc). But this problem becomes difficult when dealing with piecewise constant functions whose endpoints are \emph{continuous}, as the endpoints of each interval in the step function may be at any location. Rasterizing the step function and performing convolution is not a viable solution, as this fails in the common case of extremely narrow histogram bins.

A new algorithm is therefore required. Consider a step function $(\t, \y)$ where $(x_i, x_{i+1})$ are the endpoints of interval $i$ and $y_i$ is the value of the step function inside interval $i$. We want to convolve this step function with a rectangular pulse with radius $r$ that integrates to $1$: $[ |\t| < r ]/(2r)$ where $[\,]$ are Iverson brackets.
Observe that the convolution of a single interval $i$ of a step function with this rectangular pulse is a piecewise linear trapezoid with knots at $(x_i - r, 0), (x_i + r, y_i), (x_{i+1} - r, y_i), (x_{i+1} + r, 0)$,
and that a piecewise linear spline is the double integral of scaled delta functions located at each spline knot~\cite{heckbert1986filtering}. 
With this, and with the fact that summation commutes with integration, we can efficiently convolve a step function with a rectangular pulse as follows: We turn each endpoint $x_i$ of the step function into two signed delta functions located at $x_i - r$ and $x_i + r$ with values that are proportional to the change between adjacent $y$ values, we interleave (via sorting) those delta functions, and we then integrate those sorted delta functions twice (see Algorithm~1 of the supplement for pseudocode). With this, we can construct our anti-aliased loss function.

\myparagraph{Anti-Aliased Interlevel Loss} The proposal supervision approach in mip-NeRF 360, which we inherit, requires a loss function that takes as input a step function produced by the NeRF $(\mathbf{s}, \mathbf{w})$ and a similar step function  produced by the proposal model $(\hat{\mathbf{s}}, \hat{\mathbf{w}})$. Both of these step functions are histograms, where $\mathbf{s}$ and $\hat{\mathbf{s}}$ are vectors of endpoint locations while $\mathbf{w}$ and $\hat{\mathbf{w}}$ are vectors of weights that sum to $\leq 1$, where $w_i$ denote how visible scene content is interval $i$ of the step function. Each $s_i$ is some normalized function of true metric distance $t_i$, according to some normalization function $g(\cdot)$, which we discuss later. Note that $\mathbf{s}$ and $\hat{\mathbf{s}}$ are not the same --- the endpoints of each histogram are distinct.

To train a proposal network to bound the scene geometry predicted by the NeRF without introducing aliasing, we require a loss function that can measure the distance between $(\mathbf{s}, \mathbf{w})$ and $(\hat{\mathbf{s}}, \hat{\mathbf{w}})$ that is smooth with respect to translation along the ray, despite the challenge that the endpoints of both step functions are different. To do this, we will blur the NeRF histogram $(\mathbf{s}, \mathbf{w})$ using our previously-constructed algorithm and then resample that blurred distribution into the set of intervals of the proposal histogram $\hat{\mathbf{s}}$ to produce a new set of histogram weights $\mathbf{w}^{\hat{\mathbf{s}}}$. This procedure is described in Figure~\ref{fig:zaliasing_solution}.
After resampling our blurred NeRF weights into the space of our proposal histogram, our loss function is an element-wise function of $\mathbf{w}^{\hat{\mathbf{s}}}$ and $\hat{\mathbf{w}}$:
\begin{equation}
\textstyle \proploss\left(\mathbf{\sdistance}, \mathbf{w}, \proposal{\mathbf{\sdistance}}, \proposal{\mathbf{w}} \right) = 
\sum_{i} 
 \frac{1}{\proposal{w}_i}\operatorname{max}(0, \stopgrad(w_i^{\hat{\mathbf{s}}}) - \proposal{w_i})^2 \,.
\end{equation}
Though this loss resembles mip-NeRF 360's (a half-quadratic chi-squared loss with a stop-gradient), the blurring and resampling used to generate $\mathbf{w}^{\hat{\mathbf{s}}}$ prevents aliasing.

\begin{figure}[b]
    \centering
    \begin{tabular}{@{}c@{\,\,}c@{}}
    \includegraphics[width=0.48\linewidth]{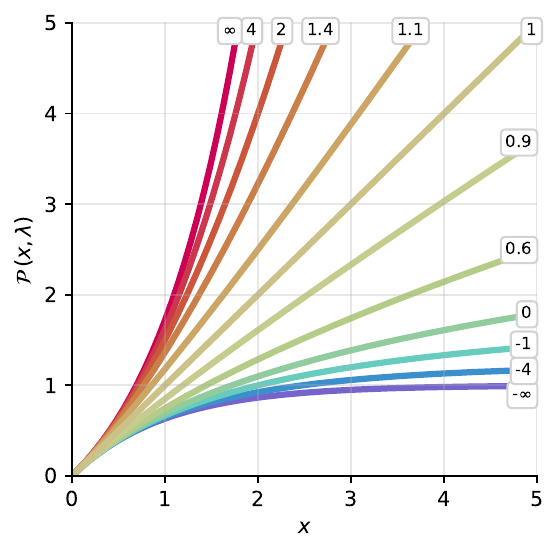} &
    \includegraphics[width=0.49\linewidth]{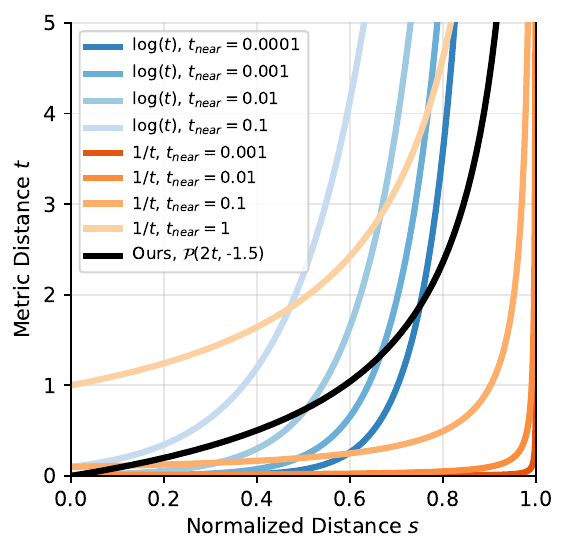}
    \end{tabular}
    \vspace{-0.07in}
    \caption{
    Many NeRF approaches require a function for converting metric distance $t \in [0, \infty)$ into normalized distance $s \in [0, 1]$. Left: Our power transformation $\powerladder(x, \power)$ lets us interpolate between common curves such as linear, logarithmic, and inverse by modifying $\lambda$, while retaining a linear-like shape near the origin. Right: This lets us construct a curve that transitions from linear to inverse/inverse-square, and supports scene content close to the camera.
    \label{fig:spacing}
    }
\end{figure}

\myparagraph{Normalizing Metric Distance} As in mip-NeRF 360, we parameterize metric distance along a ray $t \in [t_{\mathit{near}}, t_{\mathit{far}}]$ in terms of normalized distance $s \in [0, 1]$  (where $t_{\mathit{near}}$ and $t_{\mathit{far}}$ are manually-defined near and far plane distances). Rendering uses metric distance $t$, but resampling and proposal supervision use normalized distance $s$, with some function $g(\cdot)$ defining a bijection between the two. 
The interlevel loss used in mip-NeRF 360 is invariant to monotonic transformations of distance, so it is unaffected by the choice of $g(\cdot)$. However, the prefiltering in our anti-aliased loss removes this invariance, and using mip-NeRF 360's $g(\cdot)$ in our model results in catastrophic failure, so we must construct a new normalization. To do this, we construct a novel power transformation~\cite{box1964analysis,tukey77}:
\begin{equation}
\powerladder(x, \power) = 
  \frac{|\power - 1|}{\power}\left(\!\left(\!\frac{x}{|\power-1|} + 1 \!\right)^{\!\power} - 1 \!\right)\,.
\end{equation}
The slope of this function at the origin is 1, so normalized distance near the ray origin is proportional to metric distance (obviating the need to tune a non-zero near plane distance $t_{\mathit{near}}$), but far from the origin metric distance becomes curved to resemble log-distance~\cite{donerf} ($\lambda=0$) or inverse-distance~\cite{barron2022mipnerf360} ($\lambda=-1$). This lets us smoothly interpolate between different normalizations, instead of swapping in different discrete functions.
See Figure~\ref{fig:spacing} for a visualization of $\powerladder(x, \power)$ and a comparison of prior normalization approaches to ours, which is $g(x) = \powerladder(2x, -1.5)$ --- a curve that is roughly linear when $s \in [0, \sfrac{1}{2})$ but is between inverse and inverse-square when $s \in [\sfrac{1}{2}, 1]$.

\section{Results}
\label{sec:details}

Our model is implemented in JAX~\cite{jax2018github} and based on the mip-NeRF 360 codebase~\cite{multinerf2022}, with a reimplementation of iNGP's pyramid of voxel grids and hashes in place of the large MLP used by mip-NeRF 360. Our overall model architecture is identical to mip-NeRF 360 except for the anti-aliasing adjustments introduced in Sections~\ref{sec:spatial_aliasing} and \ref{sec:z_aliasing}, as well as some additional modifications that we describe here. 

Like mip-NeRF 360, we use two rounds of proposal sampling with 64 samples each, and then 32 samples in the final NeRF sampling round. Our anti-aliased interlevel loss is imposed on both rounds of proposal sampling, with a rectangular pulse width of $r=0.03$ in the first round and $r=0.003$ in the second round, and with a loss multiplier of $0.01$. We use separate proposal iNGPs and MLPs for each round of proposal sampling, and our NeRF MLP uses a much larger view-dependent branch than is used by iNGP. See the supplement for details.

One small but important modification we make to iNGP is imposing a normalized weight decay on the feature codes stored in its pyramid of grids and hashes: $\sum_\ell  \operatorname{mean}\left(\grid_\level^2 \right) $. By penalizing \emph{the sum of the mean} of squared grid/hash values at each pyramid level $\grid_\ell$ we induce very different behavior than the naive solution of penalizing the sum of all values, as coarse scales are penalized by orders of magnitude more than fine scales. This simple trick is extremely effective --- it improves performance greatly compared to no weight decay, and significantly outperforms naive weight decay. We use a loss multiplier of $0.1$ on this normalized weight decay in all experiments unless otherwise stated.


\begin{table*}[t!]
    \centering
    \resizebox{\linewidth}{!}{
    \begin{tabular}{@{}c@{\,}@{\,}l@{\,\,}|ccc|ccc|ccc|ccc|c@{}}
    & \multicolumn{1}{r|}{Scale Factor:} & \multicolumn{3}{c|}{1$\times$} & \multicolumn{3}{c|}{2$\times$} & \multicolumn{3}{c|}{4$\times$} & \multicolumn{3}{c|}{8$\times$} & \\
    & \multicolumn{1}{r|}{Error Metric:}  & \!PSNR $\uparrow$\! & \!SSIM $\uparrow$\! & \!LPIPS $\downarrow$\! & \!PSNR $\uparrow$\! & \!SSIM $\uparrow$\! & \!LPIPS $\downarrow$\! & \!PSNR $\uparrow$\! & \!SSIM $\uparrow$\! & \!LPIPS $\downarrow$\! & \!PSNR $\uparrow$\! & \!SSIM $\uparrow$\! & \!LPIPS $\downarrow$\! & Time (hrs) \\ \hline
    & Instant NGP~\cite{muller2022instant,bakedsdf}           &                   24.36 &                   0.642 &                   0.366 &                   25.23 &                   0.712 &                   0.251 &                   26.84 &                   0.809 &                   0.142 &                   28.42 &                   0.877 &                   0.092 & 0.15 \\
& mip-NeRF 360~\cite{barron2022mipnerf360, multinerf2022} &                   27.51 &                   0.779 &                   0.254 &                   29.19 &                   0.864 &                   0.136 &                   30.45 &                   0.912 &                   0.077 &                   30.86 &                   0.931 &                   0.058 & 21.86 \\
& mip-NeRF 360 + iNGP                                     &                   26.46 &                   0.773 &                   0.253 &                   27.92 &                   0.855 &                   0.141 &                   27.67 &                   0.866 &                   0.116 &                   25.58 &                   0.804 &                   0.160 & 0.31 \\
\hline
& Our Model                                               & \cellcolor{orange}28.25 & \cellcolor{orange}0.822 & \cellcolor{orange}0.198 & \cellcolor{orange}30.00 &    \cellcolor{red}0.892 & \cellcolor{orange}0.099 &    \cellcolor{red}31.57 &    \cellcolor{red}0.933 &    \cellcolor{red}0.056 &    \cellcolor{red}32.52 &    \cellcolor{red}0.954 &    \cellcolor{red}0.037 & 0.90 \\
\hline
A) & Naive Sampling                                       &                   27.93 &                   0.797 &                   0.233 &                   29.70 &                   0.880 &                   0.114 &                   29.24 &                   0.887 &                   0.094 &                   26.53 &                   0.820 &                   0.144 & 0.53 \\
B) & Naive Supersampling ($6\times$)                      &                   27.48 &                   0.803 &                   0.224 &                   29.03 &                   0.881 &                   0.109 &                   28.42 &                   0.881 &                   0.097 &                   25.97 &                   0.810 &                   0.151 & 2.54 \\
C) & Jittered                                             &                   27.91 &                   0.797 &                   0.233 &                   29.60 &                   0.879 &                   0.116 &                   29.45 &                   0.893 &                   0.090 &                   27.58 &                   0.855 &                   0.120 & 0.55 \\
D) & Jittered Supersampling ($6\times$)                   &                   27.50 &                   0.810 &                   0.212 &                   28.99 &                   0.884 &                   0.105 &                   28.91 &                   0.896 &                   0.086 &                   27.65 &                   0.870 &                   0.109 & 3.04 \\
E) & No Multisampling                                     &                   28.15 &                   0.817 &                   0.208 &                   29.87 &                   0.886 &                   0.105 & \cellcolor{yellow}31.33 &                   0.927 &                   0.061 &                   32.12 &                   0.948 &                   0.043 & 0.54 \\
F) & No Downweighting                                     &                   28.22 &                   0.818 &                   0.205 &                   29.94 &                   0.889 &                   0.102 &                   31.25 &                   0.928 &                   0.060 &                   31.67 &                   0.944 &                   0.046 & 0.88 \\
G) & No Appended Scale $\downweight$                      & \cellcolor{yellow}28.23 &                   0.820 & \cellcolor{yellow}0.200 & \cellcolor{yellow}29.98 & \cellcolor{yellow}0.890 &                   0.101 & \cellcolor{orange}31.48 & \cellcolor{yellow}0.931 & \cellcolor{orange}0.057 &                   32.19 & \cellcolor{yellow}0.951 &                   0.041 & 0.89 \\
H) & Random Multisampling                                 &                   28.09 &                   0.816 &                   0.207 &                   29.75 &                   0.886 &                   0.106 &                   31.18 &                   0.928 &                   0.061 &                   32.03 &                   0.950 &                   0.042 & 0.95 \\
I) & Unscented Multisampling                              &    \cellcolor{red}28.27 & \cellcolor{orange}0.822 & \cellcolor{orange}0.198 &    \cellcolor{red}30.03 & \cellcolor{orange}0.891 & \cellcolor{yellow}0.100 &    \cellcolor{red}31.57 &    \cellcolor{red}0.933 &    \cellcolor{red}0.056 & \cellcolor{orange}32.46 &    \cellcolor{red}0.954 & \cellcolor{orange}0.038 & 1.11 \\
J) & No New Interlevel Loss                               &                   28.12 &    \cellcolor{red}0.824 &    \cellcolor{red}0.196 &                   29.82 &    \cellcolor{red}0.892 &    \cellcolor{red}0.098 &                   31.31 & \cellcolor{orange}0.932 &    \cellcolor{red}0.056 & \cellcolor{yellow}32.23 & \cellcolor{orange}0.953 & \cellcolor{yellow}0.039 & 0.86 \\
K) & No Weight Decay                                      &                   27.34 &                   0.814 &                   0.203 &                   28.91 &                   0.881 &                   0.109 &                   30.29 &                   0.921 &                   0.067 &                   31.23 &                   0.941 &                   0.050 & 0.90 \\
L) & Un-Normalized Weight Decay                           &                   27.99 & \cellcolor{yellow}0.821 &    \cellcolor{red}0.196 &                   29.65 &                   0.889 & \cellcolor{yellow}0.100 &                   31.10 &                   0.930 & \cellcolor{yellow}0.058 &                   32.09 & \cellcolor{yellow}0.951 &                   0.040 & 0.91 \\
M) & Small View-Dependent MLP                             &                   27.41 &                   0.811 &                   0.207 &                   28.98 &                   0.882 &                   0.109 &                   30.32 &                   0.924 &                   0.065 &                   31.11 &                   0.944 &                   0.047 & 0.63  \!\!\!
    \end{tabular}
    }
    \caption{
    Performance on our multiscale version of the 360 dataset~\cite{barron2022mipnerf360}, where we train and evaluate on multiscale images, as was done in mip-NeRF~\cite{barron2021mipnerf}. Red, orange, and yellow highlights indicate the 1st, 2nd, and 3rd-best performing technique for each metric. Our model significantly outperforms our two baselines --- especially the iNGP-based one, and especially at coarse scales, where our errors are 55\%--77\% reduced. Rows A--M are ablations of our model, see the text for details.
    }
    \label{tab:360ms}
\end{table*}

\myparagraph{360 Dataset} Since our model is an extension of mip-NeRF 360, we evaluate on the ``360'' benchmark presented in that paper using the same evaluation procedure~\cite{barron2022mipnerf360}. We evaluate NeRF~\cite{mildenhall2020}, mip-NeRF~\cite{barron2021mipnerf}, NeRF++~\cite{kaizhang2020}, mip-NeRF 360, iNGP~\cite{muller2022instant} (using carefully tuned hyperparameters taken from recent work~\cite{bakedsdf}), and a baseline in which we naively combine mip-NeRF 360 and iNGP without any of this paper's contributions.
Average error metrics across all scenes are shown in Table~\ref{tab:avg_360_results} (see the supplement for per-scene metrics) and renderings of these four approaches are shown in Figure~\ref{fig:single_results}. Our model, mip-NeRF 360, and our ``mip-NeRF 360 + iNGP'' baseline were all trained on 8 NVIDIA Tesla V100-SXM2-16GB GPUs. The other baselines were trained on different accelerators (TPUs were used for NeRF, mip-NeRF, and NeRF++, and a single NVIDIA 3090 was used for iNGP) so to enable comparison their runtimes have been rescaled to approximate performance on our hardware.
The render time of our model (which is not a focus of this work) is $0.9$s, while mip-NeRF 360 takes $7.4$s and our mip-NeRF 360 + iNGP baseline takes $0.2$s.

Our model significantly outperforms mip-NeRF 360 (the previous state-of-the-art of this task) on this benchmark: we observe 11\%, 17\%, and 19\% reductions in RMSE, DSSIM, and LPIPS respectively. Additionally, our model trains $24 \times$ faster than mip-NeRF 360. Compared to iNGP, our error reduction is even more significant: 28\%, 42\%, and 37\% reductions in RMSE, DSSIM, and LPIPS, though our model is $\sim\!\!6 \times$ slower to train than iNGP. Our combined ``mip-NeRF 360 + iNGP'' baseline trains significantly faster than mip-NeRF 360 (and is nearly $3 \times$ faster to train than our model) and produces error rates comparable to mip-NeRF 360's and better than iNGP's. Our model's improvement in image quality over these baselines is clear upon visual inspection, as shown in Figure~\ref{fig:single_results} and the supplemental video.

\begin{table}[b!]
    \centering
    \resizebox{0.95\linewidth}{!}{
    \begin{tabular}{@{}l@{\,\,}|ccc|@{\,}c@{}}
    & \!PSNR $\uparrow$\! & \!SSIM $\uparrow$\! & \!LPIPS $\downarrow$\! & Time (hrs) \\ \hline
    NeRF~\cite{mildenhall2020,jaxnerf2020github}            &                   23.85 &                   0.605 &                   0.451 & 12.65 \\
mip-NeRF~\cite{barron2021mipnerf}                       &                   24.04 &                   0.616 &                   0.441 & 9.64 \\
NeRF++~\cite{kaizhang2020}                              &                   25.11 &                   0.676 &                   0.375 & 28.73 \\
Instant NGP ~\cite{muller2022instant,bakedsdf}          &                   25.68 &                   0.705 &                   0.302 & 0.15 \\
mip-NeRF 360~\cite{barron2022mipnerf360, multinerf2022} & \cellcolor{orange}27.57 & \cellcolor{orange}0.793 & \cellcolor{yellow}0.234 & 21.69 \\
mip-NeRF 360 + iNGP                                     & \cellcolor{yellow}26.43 & \cellcolor{yellow}0.786 & \cellcolor{orange}0.225 & 0.30 \\
\hline
Our Model                                               &    \cellcolor{red}28.54 &    \cellcolor{red}0.828 &    \cellcolor{red}0.189 & 0.89  \!\!\!
    \end{tabular}
    }
    \caption{
    Performance on the 360 dataset~\cite{barron2022mipnerf360}. 
    Runtimes shown in gray are approximate, see text for details.
    }
    \label{tab:avg_360_results}
\end{table}

\myparagraph{Multiscale 360 Dataset}
Though the 360 dataset contains challenging scene content, it does not measure rendering quality as a function of \emph{scale} --- because this dataset was captured by orbiting a camera around a central object at a roughly constant distance, learned models need not generalize well across different image resolutions or different distances from the central object. We therefore use a more challenging evaluation procedure, similar to the multiscale Blender dataset used by mip-NeRF~\cite{barron2021mipnerf}: we turn each image into a set of four images that have been bicubically downsampled by scale factors $[1, 2, 4, 8]$ where the downsampled images act as a proxy for additional training/test views in which the camera has been zoomed out from the center of the scene. During training we multiply the data term by the scale factor of each ray, and at test time we evaluate each scale separately. This significantly increases the difficulty of reconstruction by requiring the learned model to generalize across scales, and it causes aliasing artifacts to be highly salient, especially at coarse scales.

In Table~\ref{tab:360ms} we evaluate our model against iNGP, mip-NeRF 360, our mip-NeRF 360 + iNGP baseline, and many ablations. Though mip-NeRF 360 performs reasonably (as it can reason about scale) our model still reduces RMSE by 8\% at the finest scale and 17\% at the coarsest scale, while being $24\times$ faster. The mip-NeRF 360 + iNGP baseline, which has no mechanism for anti-aliasing or reasoning about scale, performs poorly: our RMSEs are 19\% lower at the finest scale and 55\% lower at the coarsest scale, and both DSSIM and LPIPS at the coarsest scale are 77\% lower. This improvement can be seen in Figure~\ref{fig:multi_results}. Our mip-NeRF 360 + iNGP baseline generally outperforms iNGP (except at the coarsest scale), as we might expect from Table~\ref{tab:avg_360_results}.

In Table~\ref{tab:360ms} we also include an ablation study of our model. (A) Using ``naive'' sampling with our multisampling and downweighting disabled (which corresponds to the approach used in NeRF and iNGP) significantly degrades quality, and (B) supersampling~\cite{cook1986stochastic} by a $6 \times$ factor to match our multisample budget does not improve accuracy. (C, D) Randomly jittering these naive samples within the cone being cast improves performance at coarse scales, but only slightly. Individually disabling (E) multisampling and (F) downweighting lets us measure the impact of each component; they both contribute evenly. (G) Not using $\downweight$ as a feature hurts performance slightly. Alternative multisampling approaches like (H) sampling 6 random points from each mip-NeRF Gaussian or (G) using 7 unscented transform control points from each Gaussian~\cite{julier2004unscented} as multisamples are competitive alternatives to our hexagonal pattern, but have slightly worse speeds and/or accuracies. (J) Disabling our anti-aliased interlevel loss has little effect on our single-image metrics, but causes $z$-aliasing artifacts in our video results. (K) Disabling our normalized weight decay reduces accuracy significantly, and (L) using non-normalized weight decay performs poorly. (M) Using iNGP's small view-dependent MLP decreases accuracy.

\renewcommand{\mywidth}{0.239\linewidth}
\begin{figure}[t]
    \centering
    \begin{tabular}{@{}c@{\,\,}c@{\,\,}c@{\,\,}c@{}}
        \includegraphics[width=\mywidth]{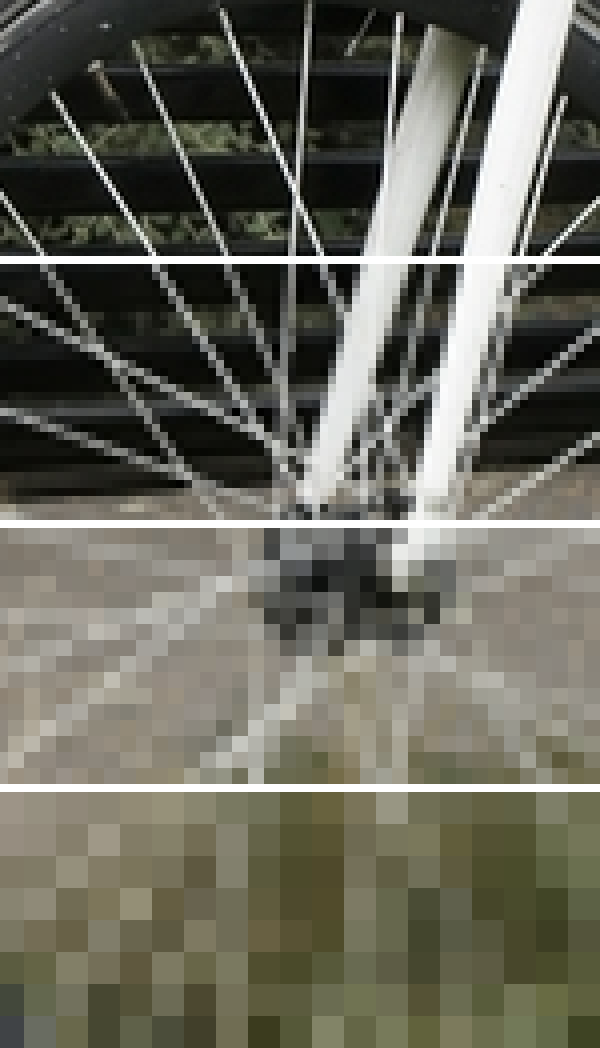} &
        \includegraphics[width=\mywidth]{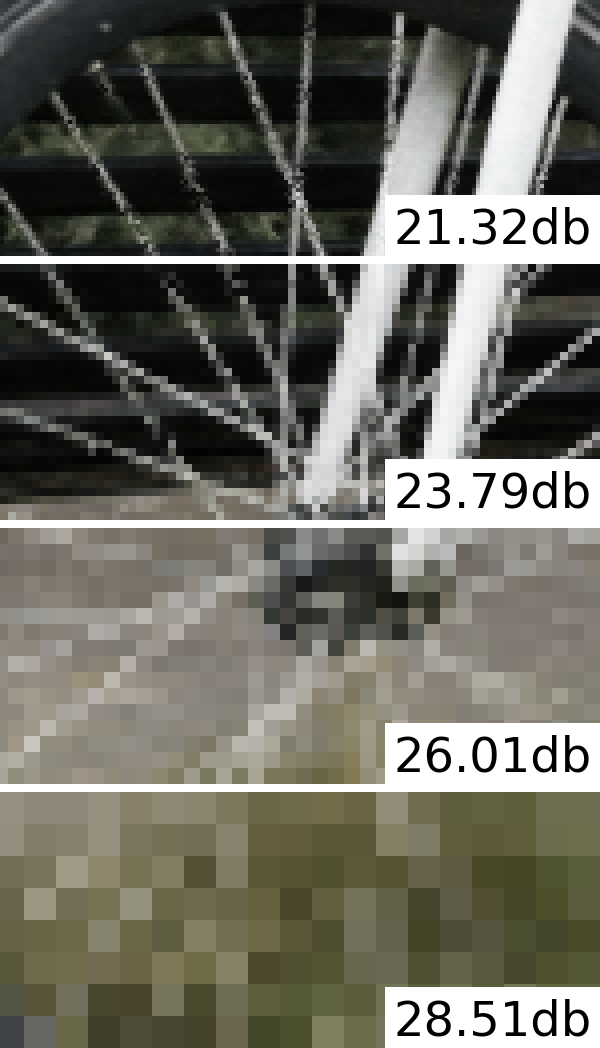} &
        \includegraphics[width=\mywidth]{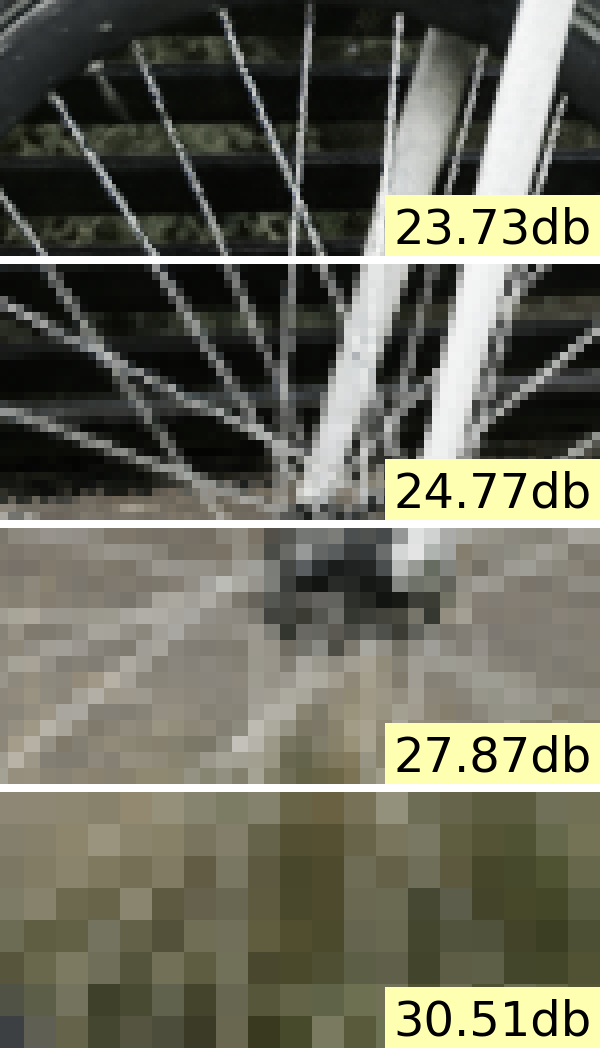} & \includegraphics[width=\mywidth]{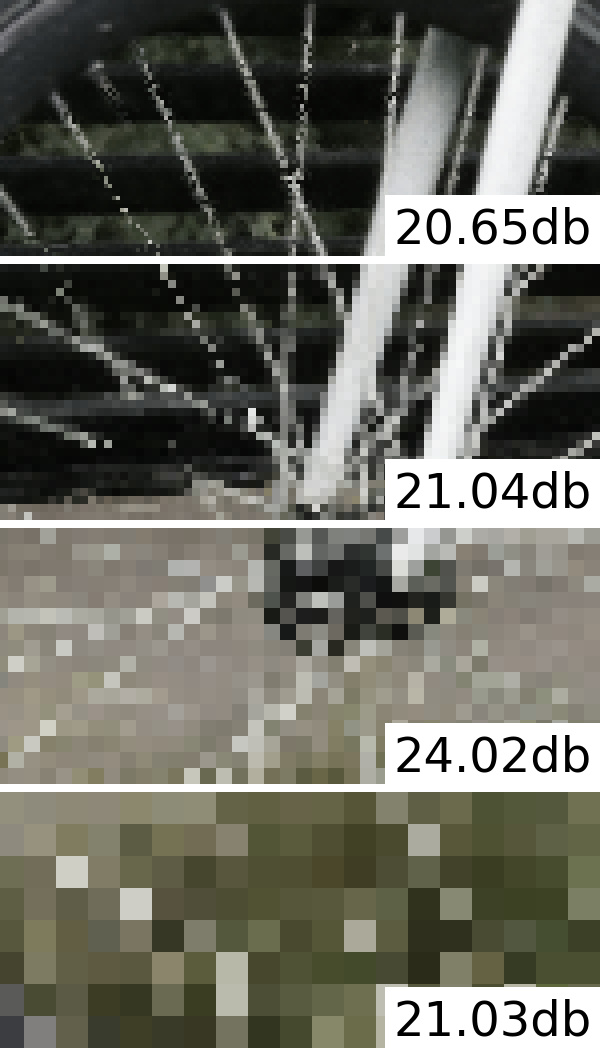} \\[-0.75ex]
        \scriptsize (a) Ground Truth & \scriptsize (b) mip-NeRF 360 & \scriptsize (c) Our Model & \! \scriptsize (d) mip360 + iNGP
    \end{tabular}
    \caption{
    A test-set view from our multiscale benchmark at (a) the four different scales that we train and test on. (b) Mip-NeRF 360 behaves reasonably across scales but is outperformed by (c) our model, especially on thin structures. (d) Our iNGP-based baseline produces extremely aliased renderings, especially at coarse scales. PSNRs are inset.
    \label{fig:multi_results}
    }
\end{figure}

\begin{figure}[t]
    \centering
    \includegraphics[trim=0 2.22in 0 0in, clip, width=\linewidth]{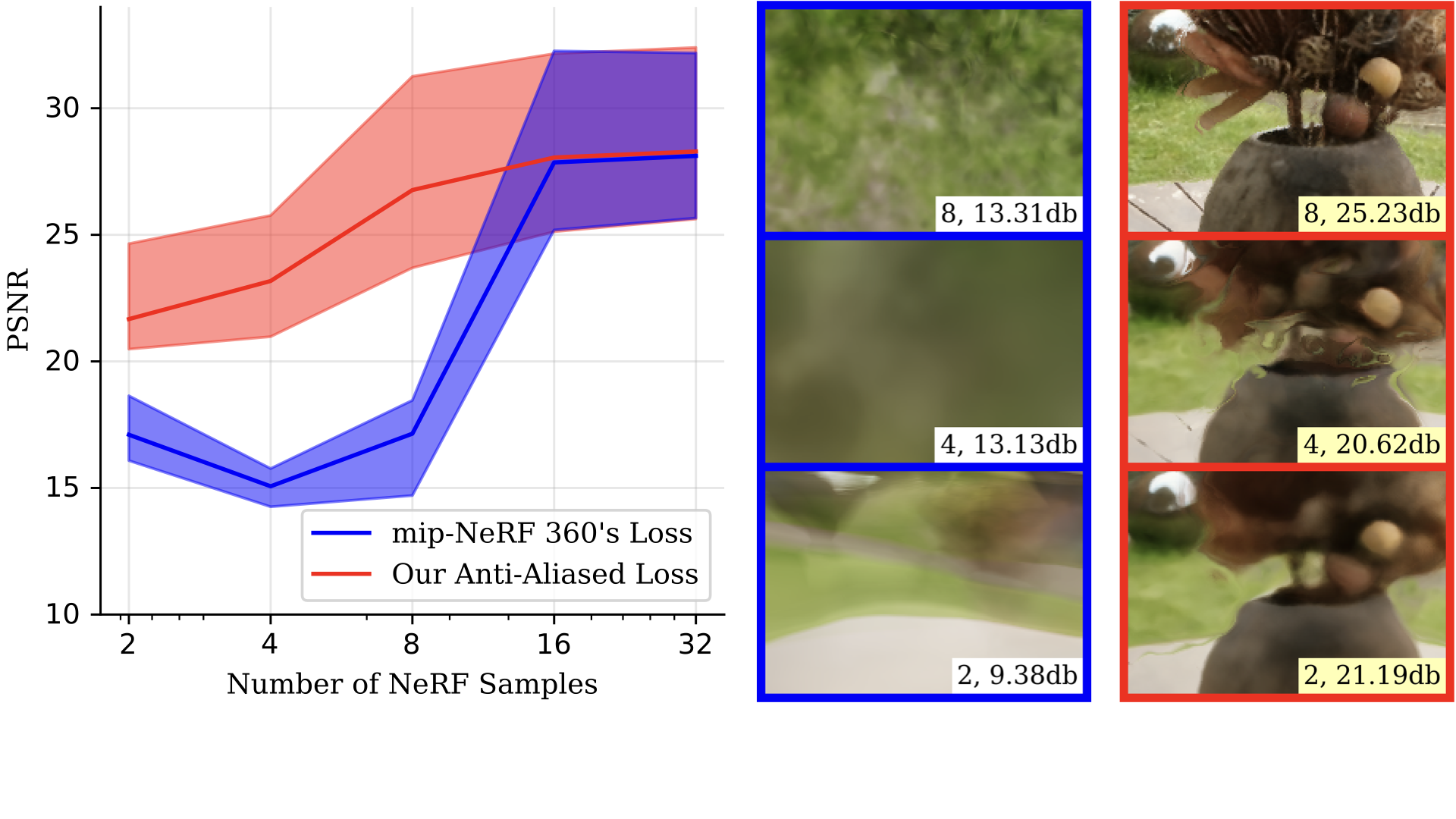}
    \caption{
    Our anti-aliased interlevel loss function (red) lets performance degrades gracefully as the number of NeRF samples is reduced from 32 to 2. Mip-NeRF 360's loss (blue) results in catastrophic failure if fewer than 16 samples are used.
    Here we plot the mean and IQR of test-set PSNR. NeRF sample counts and PSNRs are inset.
    \label{fig:z_results}
    }
\end{figure}

\myparagraph{Sample Efficiency}
Ablating our anti-aliased interlevel loss in our multiscale benchmark does not degrade quality significantly because mip-NeRF 360 (and our model, for the sake of fair comparison) uses a large number of samples\footnote{Note that here we use ``samples'' to refer to sampled sub-frusta along the ray, not the sampled multisamples \emph{within} each sub-frustum.} along each ray, which reduces $z$-aliasing. To reveal mip-NeRF 360's vulnerability to $z$-aliasing, in Figure~\ref{fig:z_results} we plot test-set PSNR for our multiscale benchmark as we reduce the number of samples along the ray (both for the NeRF and the proposal network) by factors of 2, while also doubling $r$ and the number of training iterations. Though our anti-aliased loss only slightly outperforms mip-NeRF 360's loss when the number of samples is large, as the sample count decreases the baseline loss fails catastrophically while our loss degrades gracefully.

That said, the most egregious $z$-aliasing artifact of missing scene content (shown in Figure~\ref{fig:zaliasing}) is hard to measure in small benchmarks of still images, as scene content only disappears at certain distances that the test set may not probe. See the supplemental video for demonstrations of $z$-aliasing, and for renderings from our model where that aliasing has been fixed.

\renewcommand{\mywidth}{0.49\linewidth}

\section{Conclusion}

We have presented Zip-NeRF, a model that integrates the progress made in the formerly divergent areas of scale-aware anti-aliased NeRFs and fast grid-based NeRF training.
By leveraging ideas about multisampling and prefiltering, our model is able to achieve error rates that are 8\% -- 77\% lower than prior techniques, while also training $24\times$ faster than mip-NeRF 360 (the previous state-of-the-art on our benchmarks).
We hope that the tools and analysis presented here concerning aliasing (both the spatial aliasing of NeRF's learned mapping from spatial coordinate to color and density, and $z$-aliasing of the loss function used during online distillation along each ray) enable further progress towards improving the quality, speed, and sample-efficiency of NeRF-like inverse rendering techniques.

{\small
\bibliographystyle{ieee_fullname}
\bibliography{arxiv}
}

\appendix

\section{Video Results}

This supplement includes video results for scenes from the 360 benchmark. We also present video results for new more-challenging scenes that we have captured to qualitatively demonstrate various kinds of aliasing and our model's ability to ameliorate that aliasing.

\paragraph{Affine Generative Latent Optimization}
The video results presented in mip-NeRF 360 use the appearance-embedding approach of NeRF-W~\cite{martinbrualla2020nerfw}, which assigns short GLO vectors~\cite{bojanowski2018optimizing} to each image in the training set that are concatenated to the input of the view-dependent MLP. By jointly optimizing over these embeddings during training, optimization is able to explain away view-dependent effects such as variation in exposure and lighting. On scenes with significant illumination variation we found this approach to be reasonably effective, but it is limited by the relatively small size of the view-dependent branch of our MLP compared to NeRF-W and mip-NeRF 360. We therefore use an approach similar to NeRF-W's GLO approach, but instead of optimizing over a short feature that is concatenated onto our bottleneck vector, we optimize over an affine transformation of our bottleneck vector itself. Furthermore, we express this affine transformation not just with a per-image latent embedding, but also with a small MLP that maps from a per-image latent embedding to an affine transformation. We found that optimizing over the large space of embeddings and MLP weights resulted in faster training than the GLO baseline, which allows the model to more easily explain per-image appearance variation without placing floaters in front of training cameras.

We allocate an 128-length vector for each image in the training set and use those vectors as input to a two-layer MLP with 128 hidden units whose output is two vectors (scale and shift) that are the same length as the bottleneck. The internal activation of the MLP is a ReLU, and the final activation that yields our scaling is $\exp$. We scale and shift each bottleneck by the two MLP outputs before evaluating the view-dependent MLP. This approach of optimizing an affine function of an internal activation resembles prior techniques in the literature for modulating activations to control ``style'' and appearance~\cite{huang2017adain, park2019SPADE, perez2018film}. This technique is only used to produce our video results and is not used for the experiments mentioned in the paper, as it does not improve quantitative performance on our metrics.

\section{Multisampling Pattern Derivation}

The hexagonal multisampling pattern presented in the paper was constructed so as to satisfy several criteria:
\begin{compactenum}
    \item Samples should be uniformly distributed along the ray, to ensure good coverage along the ray.
    \item Samples should be uniformly distributed in terms of angles around the ray.
    \item The sample mean and covariance of the set of samples should match the analytical mean and covariance of the conical frustum.
    \item The number of points should be as small as possible, for the sake of efficiency.
\end{compactenum}
We additionally chose to always distribute samples at a distance from the ray that is proportional to the radius of the cone at whatever $t$ value the sample is located. This simplifies the analysis of our pattern, though it does mean that none of our samples are placed exactly along the ray, which may be contrary to the reader's expectations. Experimentally, we found little value in adding additional multisamples exactly along the ray, which is consistent with the performance of the unscented transform baseline (which places multiple points along the ray).

Before constructing our conical frustum-shaped multisampling pattern, we can simplify our analysis by first constructing an $n$-point multisampling pattern for a cylinder. Here is a set of $n$ coordinates that, for a carefully chosen $\boldsymbol{\theta}$ and $n$, has a mean of $\vec{0}$ and a covariance of $\mathrm{I}_3$:
\begin{gather}
\left\{ {\begin{bmatrix} \cos(\theta_j)/\sqrt{3} \\ \sin(\theta_j)/\sqrt{3} \\ \frac{j - (n-1)/2}{\sqrt{n(n^2-1)/12}} \end{bmatrix} \,\Bigg|\, \begin{matrix} j = 0, 1, \ldots, n \end{matrix} } \right\}\,.
\end{gather}
Perhaps surprisingly, for small values of $n$ and assuming uniformly-distributed values of $\theta$, this zero-mean and identity-covariance property appears to only hold for $n=6$ and two specific choices of $\boldsymbol{\theta}$:
\begin{gather}
  \boldsymbol{\theta}_1 = \left[\, 0, \sfrac{2\pi}{3}, \sfrac{4\pi}{3}, \sfrac{3\pi}{3}, \sfrac{5\pi}{3}, \sfrac{\pi}{3} \,\right]\,, \\
  \boldsymbol{\theta}_2 = \left[\, 0, \sfrac{3\pi}{3}, \sfrac{2\pi}{3}, \sfrac{5\pi}{3}, \sfrac{4\pi}{3}, \sfrac{\pi}{3} \,\right]\,.
\end{gather}
Because our sampling pattern is rotationally symmetric around $\theta$ and bilaterally symmetric around the $z=0$ plane of the cylinder, rotating or mirroring the coordinates corresponding to these two choices of $\boldsymbol{\theta}$ preserves their zero-mean and identity-covariance. We use $\boldsymbol{\theta}_1$ in our work, because $\boldsymbol{\theta}_2$ exhibits potentially-undesirable higher-order correlation between adjacent angles (note that $\boldsymbol{\theta}_2$ consists of three pairs of adjacent angles, while $\boldsymbol{\theta}_1$ has only two adjacent angles that are nearby).

With this cylindrical multisampling pattern that satisfies our requirements, we can then warp these samples into the shape of a conical frustum, while also shifting and scaling the coordinates such that they match the means and covariances derived in mip-NeRF~\cite{barron2021mipnerf}. This yields the formulas shown in Equations 2 and 3 in the main paper. This warping results in a slight mismatch between the sample covariance of our multisample coordinates and the covariance of the frustum: the full covariance matrices are not necessarily identical. However, the variance along the ray and the total variance perpendicular to the ray are both equal to that of the conical frustum, and the mismatch between full covariances goes to zero as $t \gg \baseradius$ (which is generally the case for real image data, where $\baseradius$ is usually small).

\section{Scale Featurization}

Along with features $\mathbf{f}_\ell$ we also average and concatenate a featurized version of $\{ \downweight_{j, \ell}\}$ for use as input to our MLP:
\begin{equation}
(2 \cdot \operatorname{mean}_j\left(\downweight_{j, \ell}\right) - 1)\sqrt{\initval^2 + \stopgrad(\operatorname{mean}(\grid_\ell^2))}\,,
\end{equation}
where $\stopgrad$ is a stop-gradient operator and $\initval$ is the magnitude used to initialize each $\grid_\level$.
This feature takes $\downweight_j$ (shifted and scaled to $[-1, 1]$) and scales it by the standard deviation of the values in $\grid_\ell$ (padded slightly using $\initval$, which guards against the case where a $\grid_\ell$'s values shrink to zero during training). This scaling ensures that our featurized scales have roughly the same magnitude features as the features themselves, regardless of how the features may grow or shrink during training. The stop-gradient prevents optimization from indirectly modifying this scale-feature by changing the values in $\grid_\ell$.

When computing the downweighting factor $\downweight$, for the sake of speed we use an approximation for $\operatorname{erf}(x)$:
\begin{equation}
\operatorname{erf}(x) \approx \operatorname{sign}(x) \sqrt{1 - \exp\left(-(\sfrac{4}{\pi})x^2\right)}\,.
\end{equation}
This has no discernible impact on quality and a very marginal impact on performance.

\section{Spatial Contraction}

Mip-NeRF 360~\cite{barron2022mipnerf360} parameterizes unbounded scenes with bounded coordinates using a spatial contraction:
\begin{equation}
\contract(\mathbf{x})  = \begin{cases}
\mathbf{x} & \norm{\mathbf{x}} \leq 1\\
\left(2  - \frac{1}{\norm{\mathbf{x}}}\right)\left(\frac{\mathbf{x}}{\norm{\mathbf{x}}}\right) & \norm{\mathbf{x}} > 1\,.\label{eq:contract}
\end{cases}
\end{equation}
This maps values in $[-\infty, \infty]^d$ to $[-2, 2]^d$ such that resolution in the contracted domain is proportional to what is required by perspective projection --- scene content near the origin is allocated significant model capacity, but distant scene content is allocated model capacity that is roughly proportional to disparity (inverse distance).

Given our multisampled isotropic Gaussians $\{\mathbf{x}_j, \sigma_j \}$, we need a way to efficiently apply a scene contraction. Contracting the means of the Gaussians simply requires evaluating each $\contract(\mathbf{x}_j)$, but contracting the scale is non-trivial, and the approach provided by mip-NeRF 360~\cite{barron2022mipnerf360} for contracting a \emph{multivariate} Gaussian is needlessly expressive and expensive for our purposes. Instead, we linearize the contraction around each $\mathbf{x}_j$ to produce $\jacobian_\contract(\mathbf{x}_j)$, the Jacobian of the contraction at $\mathbf{x}_j$, and we produce an isotropic scale in the contracted space by computing the geometric mean of the eigenvalues of $\jacobian_\contract(\mathbf{x}_j)$. This is the same as computing the determinant of the absolute value of $\jacobian_\contract(\mathbf{x}_j)$ and taking its $d$th root ($d=3$ in our case, as our coordinates are 3D):
\begin{equation}
    \contract(\isostd_j) = \isostd_j \left| \operatorname{det}\left(\jacobian_\contract(\mathbf{x}_j)\right) \right|^{\sfrac{1}{d}}\,.
\end{equation}
This is equivalent to using the approach in mip-NeRF 360~\cite{barron2022mipnerf360} to apply a contraction to a multivariate Gaussian with a covariance matrix of $\sigma^2 I_d$ and then identifying the isotropic Gaussian with the same generalized variance as the contracted multivariate Gaussian, but requires significantly less compute. 
This can be accelerated further by deriving a closed-form solution for our specific contraction:
\begin{equation}
    \left| \operatorname{det}\left(\jacobian_\contract(\mathbf{x}_j)\right) \right|^{\sfrac{1}{3}} = \left(\frac{\sqrt[3]{2 \operatorname{max}(1, \norm{\mathbf{x}_j}) - 1}}{\operatorname{max}(1, \norm{\mathbf{x}_j})}\right)^2
\end{equation}

\section{Blurring A Step Function}

Algorithm~\ref{alg:blur} contains pseudocode for the algorithm described in the paper for convolving a step function with a rectangular pulse to yield a piecewise linear spline. This code is valid JAX/Numpy code except that we have overloaded \texttt{sort()} to include the behavior of \texttt{argsort()}.

\begin{algorithm}[h!]
\caption{$\tp, \yp = \operatorname{blur\_stepfun}(\t, \y, r)$}\label{alg:blurstepfun}
\begin{algorithmic}
\State $\tp, \, \idx = \operatorname{sort}(\operatorname{concatenate}([\t - r, \t + r]))$
\State $\dy = ([\y, 0] - [0, \y]) / (2r)$
\State $\dyp = \operatorname{concatenate}([\dy, -\dy])[\idx[:\!-1]]$
\State $\yp = [0, \operatorname{cumsum}((\tp[1\!:] - \tp[:\!-1]) \operatorname{cumsum}(\dyp))]$
\end{algorithmic}
\label{alg:blur}
\end{algorithm}

\section{Power Transformation Details}

Here we expand upon the power transformation $\powerladder(x, \power)$ presented in the paper. First, let's expand upon its definition to include its two removable singularities and its limits as $\lambda$ approaches $\pm \infty$:
\begin{gather}
\powerladder(x, \power) = \begin{cases}
  x & \power = 1 \\
  \log\left(1 + x\right)  & \power = 0 \\
  e^{x} - 1 & \power = \infty \\
  1 - e^{-x} & \power = -\infty \\
  \frac{|\power - 1|}{\power}\left(\!\left(\!\frac{x}{|\power-1|} + 1 \!\right)^\power - 1 \!\right) & \text{otherwise}
\end{cases}
\end{gather}
Note that the $\lambda = -\infty, 0, +\infty$ cases can and should be implemented using the \texttt{log1p()} and \texttt{expm1()} operations that are standard in numerical computing and deep learning libraries.
As discussed, the slope of this function is 1 near the origin, but further from the origin $\lambda$ can be tuned to describe a wide family of shapes: exponential, squared, logarithmic, inverse, inverse-square, and (negative) inverse-exponential.
Scaling the $x$ input to $\powerladder$ lets us control the effective range of inputs where $\powerladder(x, \lambda)$ is approximately linear.
The second derivative of $\powerladder$ at the origin is $\pm 1$, depending on if $\power$ is more or less than 1, and the output of $\powerladder$ is bounded by $\frac{\power - 1}{\power}$ when $\power < 0$.

This power transformation relates to the general robust loss $\rho(x, \alpha, c)$~\cite{BarronCVPR2019}, as $\rho$ can be written as  $\powerladder$ applied to squared error: $\rho(x, \power, 1) = \powerladder(\sfrac{x^2}{2}, \sfrac{\power}{2})$. $\powerladder$ also resembles a reparameterized Yeo–Johnson transformation~\cite{yeo2000new}
but altered such that the transform always resembles a straight line near the origin (the second derivative of the Yeo-Johnson transformation at the origin is unbounded).

\begin{figure}[t]
    \centering
    \includegraphics[trim=0 0in 4in 0in, clip, width=\linewidth]{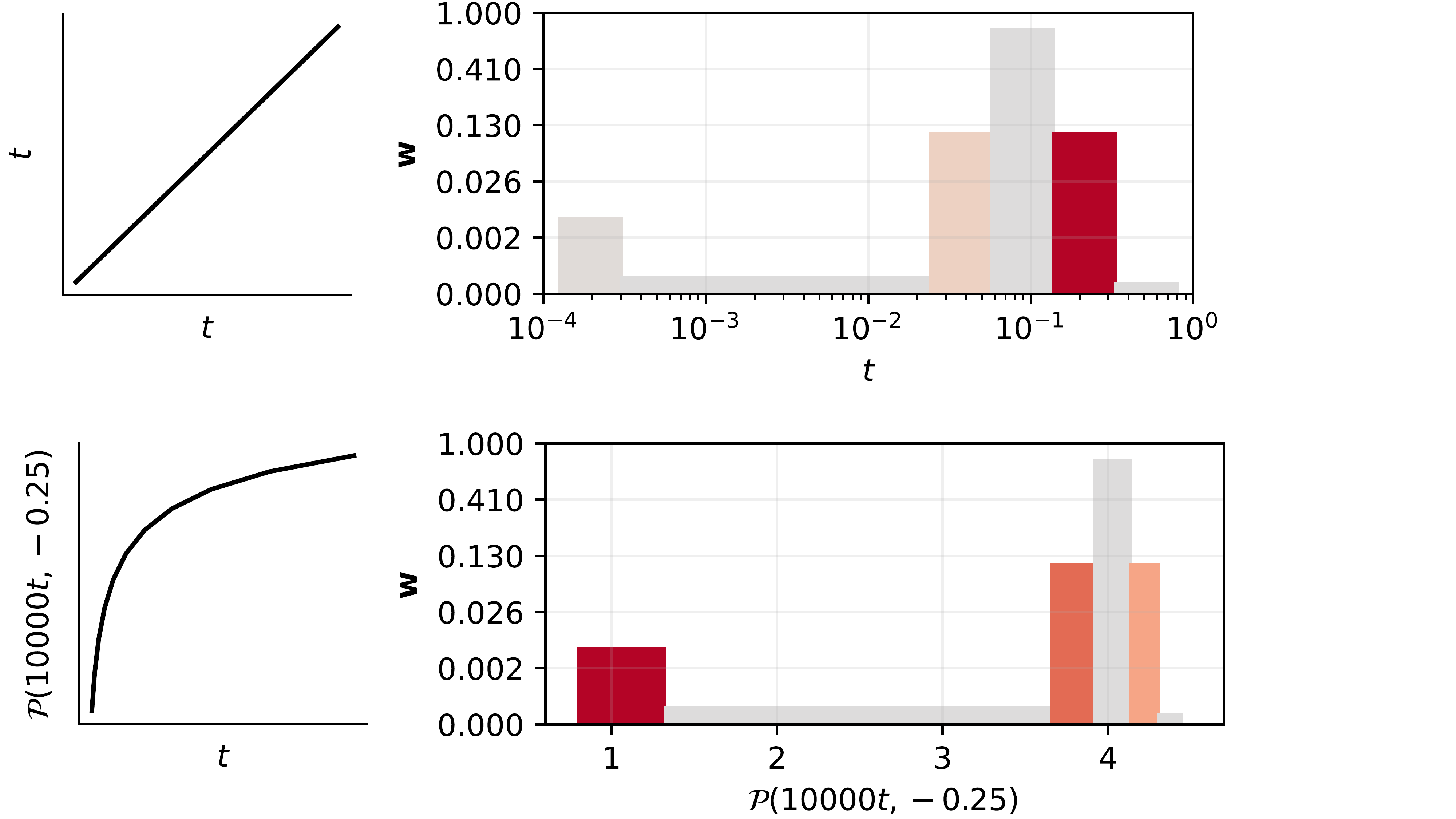}
    \caption{
    The behavior of mip-NeRF 360's distortion loss can be significantly modified by applying a curve to metric distance, which we do with our power transformation. Top: using a linear curve (\ie, using metric distance $t$ itself) results in a distortion loss that heavily penalizes distant scene content but ignores ``floaters'' close to the camera (the single large histogram bin near $t=0$), as can be seen by visualizing the gradient magnitude of distortion loss as a heat-map over the NeRF histogram shown here. Bottom: Curving metric distance with a tuned power transformation $\powerladder(10000 x, -0.25)$) before computing distortion loss causes distortion loss to correctly penalize histogram bins near the camera.
    \label{fig:distortion}
    }
\end{figure}

\paragraph{Distortion Loss}
As discussed in the paper, our power transformation is used to curve metric distance into a normalized space where resampling and interlevel supervision can be performed effectively. We also use this transformation to curve metric distance within the distortion loss used in mip-NeRF 360.
By tuning a transformation to have a steep gradient near the origin that tapers off into something resembling log-distance, we obtain a distortion loss that more aggressively penalizes ``floaters'' near the camera, which significantly improves the quality of videos rendered from our model. This does not significantly improve test-set metrics on the 360 dataset, as floaters do not tend to contribute much to error metrics computed on still images. In all of our experiments we set our model's multiplier on distortion loss to $0.005$.
See Figure~\ref{fig:distortion} for a visualization of our curve and the impact it has on distortion loss.

\section{Model Details}

Our model, our ``mip-NeRF 360 + iNGP'' baseline, and all ablations (unless otherwise stated) were trained for 25k iterations using a batch size of $2^{16}$.
We use the Adam optimizer~\cite{adam} with $\beta_1 = 0.9$, $\beta_2=0.99$, $\epsilon=10^{-15}$, and we decay our learning rate logarithmically from $10^{-2}$ to $10^{-3}$ over training. We use no gradient clipping, and we use an aggressive warm-up for our learning rate: for the first 5k iterations we scale our learning rate by a multiplier that is cosine-decayed from $10^{-8}$ to $1$. 

In our iNGP hierarchy of grids and hashes, we use 10 grid scales that are spaced by powers of 2 from $16$ to $8192$, and we use 4 channels per level. We found this to work slightly better and faster than iNGP's similar approach of spacing scales by $\sqrt{2}$ and having 2 channels per level. Grid sizes $n_\ell$ that are greater than $128$ are parameterized identically to iNGP using a hash of size $128^3$.

We inherit the proposal sampling procedure used by mip-NeRF 360: two rounds of proposal sampling where we bound scene geometry and recursively generate new sample intervals, and one final NeRF round in which we render that final set of intervals into an image. We use a distinct NGP and MLP for each round of sampling, as this improves performance slightly. Because high-frequency content is largely irrelevant for earlier rounds of proposal sampling, we truncate the hierarchy of grids of each proposal NGP at maximum grid sizes of $512$ and $2048$. We additionally only use a single channel of features for each proposal NGP, as (unlike the NeRF NGP) these models only need to predict density, not density and color.

We found that small view-dependent MLP used by iNGP to be a significant bottleneck to performance, as it limits the model's ability to express and recover complicated view-dependent effects. This not only reduces rendering quality on shiny surfaces, but also causes more ``floaters'' by encouraging optimization to explain away view-dependent effects with small floaters in front of the camera.
We therefore use a larger model: We have a view-dependent bottleneck size of 256, and we process those bottleneck vectors with a 3-layer MLP with 256 hidden units and a skip connection from the bottleneck to the second layer.

\paragraph{Ablations}
In our non-normalized weight decay ablation, we use weight decay with a multiplier of $10^{-9}$, though we found performance to be largely insensitive to what multiplier is used. Our ``Naive Supersampling ($6\times$)'' and ``Jittered Supersampling ($6\times$)'' ablations in the paper caused training to run out of memory, so those experiments use half of the batch size used in other experiments and are trained for twice as many iterations.

\section{Results}

An per-scene version of our single-scale results on the mip-NeRF 360 dataset can be found in Table~\ref{tab:scene_360_results}, and multiscale per-scene results are in Table~\ref{tab:scene_360_results_multi}. We also include per-scene and average results for the Blender dataset~\cite{mildenhall2020} in Table~\ref{tab:blender_results}. For these Blender results, we use the same model as presented in the paper, with the following changes: we set the ray near and far plane distances to 2 and 6 respectively, we use a linear (no-op) curve when spacing sample intervals along each ray, we assume a white background color, and we increase our weight decay multiplier to $10$.

\begin{table*}[b!]
    \centering
    \begin{tabular}{|l||ccccc|cccc|}
    \hline
    \bf{PSNR} & \multicolumn{5}{c|}{Outdoor} & \multicolumn{4}{c|}{Indoor} \\
     & \textit{bicycle} & \textit{flowers} & \textit{garden} & \textit{stump} & \textit{treehill} & \textit{room} & \textit{counter} & \textit{kitchen} & \textit{bonsai} \\\hline
NeRF~\cite{mildenhall2020,jaxnerf2020github}            &                   21.76 &                   19.40 &                   23.11 &                   21.73 &                   21.28 &                   28.56 &                   25.67 &                   26.31 &                   26.81 \\
mip-NeRF~\cite{barron2021mipnerf}                       &                   21.69 &                   19.31 &                   23.16 &                   23.10 &                   21.21 &                   28.73 &                   25.59 &                   26.47 &                   27.13 \\
NeRF++~\cite{kaizhang2020}                              &                   22.64 &                   20.31 &                   24.32 &                   24.34 &                   22.20 &                   28.87 & \cellcolor{yellow}26.38 &                   27.80 &                   29.15 \\
Instant NGP ~\cite{muller2022instant,bakedsdf}          &                   22.79 &                   19.19 &                   25.26 &                   24.80 &                   22.46 &                   30.31 &                   26.21 &                   29.00 &                   31.08 \\
mip-NeRF 360~\cite{barron2022mipnerf360, multinerf2022} & \cellcolor{yellow}24.40 & \cellcolor{yellow}21.64 & \cellcolor{yellow}26.94 & \cellcolor{orange}26.36 & \cellcolor{yellow}22.81 & \cellcolor{orange}31.40 &    \cellcolor{red}29.44 & \cellcolor{orange}32.02 & \cellcolor{orange}33.11 \\
mip-NeRF 360 + iNGP                                     & \cellcolor{orange}24.51 & \cellcolor{orange}21.82 & \cellcolor{orange}27.05 & \cellcolor{yellow}25.08 & \cellcolor{orange}23.01 & \cellcolor{yellow}31.07 &                   24.01 & \cellcolor{yellow}30.18 & \cellcolor{yellow}31.12 \\
\hline
Our Model                                               &    \cellcolor{red}25.80 &    \cellcolor{red}22.40 &    \cellcolor{red}28.20 &    \cellcolor{red}27.55 &    \cellcolor{red}23.89 &    \cellcolor{red}32.65 & \cellcolor{orange}29.38 &    \cellcolor{red}32.50 &    \cellcolor{red}34.46
  \!\!\! \\
    \hline
    \multicolumn{10}{c}{} \\
    \hline
    \bf{SSIM} & \multicolumn{5}{c|}{Outdoor} & \multicolumn{4}{c|}{Indoor} \\
     & \textit{bicycle} & \textit{flowers} & \textit{garden} & \textit{stump} & \textit{treehill} & \textit{room} & \textit{counter} & \textit{kitchen} & \textit{bonsai} \\\hline
NeRF~\cite{mildenhall2020,jaxnerf2020github}            &                   0.455 &                   0.376 &                   0.546 &                   0.453 &                   0.459 &                   0.843 &                   0.775 &                   0.749 &                   0.792 \\
mip-NeRF~\cite{barron2021mipnerf}                       &                   0.454 &                   0.373 &                   0.543 &                   0.517 &                   0.466 &                   0.851 &                   0.779 &                   0.745 &                   0.818 \\
NeRF++~\cite{kaizhang2020}                              &                   0.526 &                   0.453 &                   0.635 &                   0.594 &                   0.530 &                   0.852 &                   0.802 &                   0.816 &                   0.876 \\
Instant NGP ~\cite{muller2022instant,bakedsdf}          &                   0.540 &                   0.378 &                   0.709 &                   0.654 &                   0.547 &                   0.893 & \cellcolor{yellow}0.845 &                   0.857 &                   0.924 \\
mip-NeRF 360~\cite{barron2022mipnerf360, multinerf2022} & \cellcolor{orange}0.693 & \cellcolor{yellow}0.583 & \cellcolor{yellow}0.816 & \cellcolor{orange}0.746 & \cellcolor{yellow}0.632 & \cellcolor{orange}0.913 & \cellcolor{orange}0.895 & \cellcolor{orange}0.920 & \cellcolor{orange}0.939 \\
mip-NeRF 360 + iNGP                                     & \cellcolor{yellow}0.692 & \cellcolor{orange}0.615 & \cellcolor{orange}0.840 & \cellcolor{yellow}0.720 & \cellcolor{orange}0.633 & \cellcolor{yellow}0.911 &                   0.821 & \cellcolor{yellow}0.910 & \cellcolor{yellow}0.930 \\
\hline
Our Model                                               &    \cellcolor{red}0.769 &    \cellcolor{red}0.642 &    \cellcolor{red}0.860 &    \cellcolor{red}0.800 &    \cellcolor{red}0.681 &    \cellcolor{red}0.925 &    \cellcolor{red}0.902 &    \cellcolor{red}0.928 &    \cellcolor{red}0.949
  \!\!\! \\
    \hline
    \multicolumn{10}{c}{} \\
    \hline
    \bf{LPIPS} & \multicolumn{5}{c|}{Outdoor} & \multicolumn{4}{c|}{Indoor} \\
     & \textit{bicycle} & \textit{flowers} & \textit{garden} & \textit{stump} & \textit{treehill} & \textit{room} & \textit{counter} & \textit{kitchen} & \textit{bonsai} \\\hline
NeRF~\cite{mildenhall2020,jaxnerf2020github}            &                   0.536 &                   0.529 &                   0.415 &                   0.551 &                   0.546 &                   0.353 &                   0.394 &                   0.335 &                   0.398 \\
mip-NeRF~\cite{barron2021mipnerf}                       &                   0.541 &                   0.535 &                   0.422 &                   0.490 &                   0.538 &                   0.346 &                   0.390 &                   0.336 &                   0.370 \\
NeRF++~\cite{kaizhang2020}                              &                   0.455 &                   0.466 &                   0.331 &                   0.416 &                   0.466 &                   0.335 &                   0.351 &                   0.260 &                   0.291 \\
Instant NGP ~\cite{muller2022instant,bakedsdf}          &                   0.398 &                   0.441 &                   0.255 &                   0.339 &                   0.420 &                   0.242 & \cellcolor{yellow}0.255 &                   0.170 &                   0.198 \\
mip-NeRF 360~\cite{barron2022mipnerf360, multinerf2022} & \cellcolor{yellow}0.289 & \cellcolor{yellow}0.345 & \cellcolor{yellow}0.164 & \cellcolor{orange}0.254 & \cellcolor{yellow}0.338 & \cellcolor{yellow}0.211 & \cellcolor{orange}0.203 & \cellcolor{orange}0.126 & \cellcolor{yellow}0.177 \\
mip-NeRF 360 + iNGP                                     & \cellcolor{orange}0.272 & \cellcolor{orange}0.305 & \cellcolor{orange}0.134 & \cellcolor{yellow}0.256 & \cellcolor{orange}0.298 & \cellcolor{orange}0.198 &                   0.259 & \cellcolor{yellow}0.129 &    \cellcolor{red}0.171 \\
\hline
Our Model                                               &    \cellcolor{red}0.208 &    \cellcolor{red}0.273 &    \cellcolor{red}0.118 &    \cellcolor{red}0.193 &    \cellcolor{red}0.242 &    \cellcolor{red}0.196 &    \cellcolor{red}0.185 &    \cellcolor{red}0.116 & \cellcolor{orange}0.173
  \!\!\! \\
    \hline
    \end{tabular}
    \caption{
    Single-scale per-scene performance on the dataset of ``360'' indoor and outdoor scenes from mip-NeRF 360 \cite{barron2022mipnerf360}.
    }
    \label{tab:scene_360_results}
\end{table*}

\begin{table*}[b!]
    \centering
    \resizebox{\linewidth}{!}{
    \begin{tabular}{|l||@{\,\,\,}c@{\,\,\,}c@{\,\,\,}c@{\,\,\,}c@{\,\,\,}|@{\,\,\,}c@{\,\,\,}c@{\,\,\,}c@{\,\,\,}c@{\,\,\,}|@{\,\,\,}c@{\,\,\,}c@{\,\,\,}c@{\,\,\,}c@{\,\,\,}|@{\,\,\,}c@{\,\,\,}c@{\,\,\,}c@{\,\,\,}c@{\,\,\,}|@{\,\,\,}c@{\,\,\,}c@{\,\,\,}c@{\,\,\,}c@{\,\,\,}|}
    \multicolumn{21}{c}{Outdoors} \\
    \hline
    \bf{PSNR} 
    & \multicolumn{4}{c@{\,\,\,}|@{\,\,\,}}{\textit{bicycle}} 
    & \multicolumn{4}{c@{\,\,\,}|@{\,\,\,}}{\textit{flowers}}
    & \multicolumn{4}{c@{\,\,\,}|@{\,\,\,}}{\textit{garden}}
    & \multicolumn{4}{c@{\,\,\,}|@{\,\,\,}}{\textit{stump}}
    & \multicolumn{4}{c@{\,\,\,}|}{\textit{treehill}}  \\
    & $1\times$ & $2\times$ & $4\times$ & $8\times$ & $1\times$ & $2\times$ & $4\times$ & $8\times$ & $1\times$ & $2\times$ & $4\times$ & $8\times$ & $1\times$ & $2\times$ & $4\times$ & $8\times$ & $1\times$ & $2\times$ & $4\times$ & $8\times$ \\
    \hline
    mip-NeRF 360~\cite{barron2022mipnerf360, multinerf2022} & 24.51 & 26.93 & 28.53 & 29.24 & 21.64 & 23.90 & 26.01 & 27.35 & 26.71 & 29.59 & 31.35 & 32.52 & 26.27 & 27.68 & 28.82 & 29.27 & 22.93 & 24.63 & 26.06 & 27.12 \\
mip-NeRF 360 + iNGP                                     & 24.61 & 26.98 & 26.69 & 24.50 & 21.93 & 24.14 & 24.90 & 23.19 & 26.48 & 29.06 & 27.54 & 24.85 & 26.41 & 27.63 & 27.62 & 25.64 & 23.19 & 24.86 & 25.55 & 24.81 \\
\hline
Our Model                                               & \textbf{25.57} & \textbf{28.25} & \textbf{30.20} & \textbf{31.37} & \textbf{22.37} & \textbf{24.91} & \textbf{27.51} & \textbf{29.50} & \textbf{27.71} & \textbf{30.53} & \textbf{32.60} & \textbf{33.83} & \textbf{27.17} & \textbf{28.62} & \textbf{30.30} & \textbf{31.73} & \textbf{23.63} & \textbf{25.47} & \textbf{27.27} & \textbf{28.84}
  \!\!\! \\
    \hline
    \multicolumn{10}{c}{} \\
    \hline
    \bf{SSIM} & \multicolumn{4}{c@{\,\,\,}|@{\,\,\,}}{\textit{bicycle}} 
    & \multicolumn{4}{c@{\,\,\,}|@{\,\,\,}}{\textit{flowers}}
    & \multicolumn{4}{c@{\,\,\,}|@{\,\,\,}}{\textit{garden}}
    & \multicolumn{4}{c@{\,\,\,}|@{\,\,\,}}{\textit{stump}}
    & \multicolumn{4}{c@{\,\,\,}|}{\textit{treehill}}  \\
    & $1\times$ & $2\times$ & $4\times$ & $8\times$ & $1\times$ & $2\times$ & $4\times$ & $8\times$ & $1\times$ & $2\times$ & $4\times$ & $8\times$ & $1\times$ & $2\times$ & $4\times$ & $8\times$ & $1\times$ & $2\times$ & $4\times$ & $8\times$ \\ \hline
    mip-NeRF 360~\cite{barron2022mipnerf360, multinerf2022} & 0.666 & 0.815 & 0.890 & 0.912 & 0.567 & 0.727 & 0.834 & 0.881 & 0.791 & 0.903 & 0.939 & 0.959 & 0.726 & 0.819 & 0.874 & 0.882 & 0.615 & 0.748 & 0.839 & 0.893 \\
mip-NeRF 360 + iNGP                                     & 0.673 & 0.825 & 0.857 & 0.773 & 0.592 & 0.742 & 0.805 & 0.763 & 0.786 & 0.904 & 0.864 & 0.767 & 0.748 & 0.830 & 0.849 & 0.770 & 0.616 & 0.736 & 0.785 & 0.762 \\
\hline
Our Model                                               & \textbf{0.758} & \textbf{0.872} & \textbf{0.926} & \textbf{0.948} & \textbf{0.635} & \textbf{0.774} & \textbf{0.864} & \textbf{0.914} & \textbf{0.850} & \textbf{0.929} & \textbf{0.960} & \textbf{0.974} & \textbf{0.791} & \textbf{0.865} & \textbf{0.914} & \textbf{0.939} & \textbf{0.671} & \textbf{0.780} & \textbf{0.865} & \textbf{0.922}
  \!\!\! \\
    \hline
    \multicolumn{10}{c}{} \\
    \hline
    \bf{LPIPS} & \multicolumn{4}{c@{\,\,\,}|@{\,\,\,}}{\textit{bicycle}} 
    & \multicolumn{4}{c@{\,\,\,}|@{\,\,\,}}{\textit{flowers}}
    & \multicolumn{4}{c@{\,\,\,}|@{\,\,\,}}{\textit{garden}}
    & \multicolumn{4}{c@{\,\,\,}|@{\,\,\,}}{\textit{stump}}
    & \multicolumn{4}{c@{\,\,\,}|}{\textit{treehill}}  \\
    & $1\times$ & $2\times$ & $4\times$ & $8\times$ & $1\times$ & $2\times$ & $4\times$ & $8\times$ & $1\times$ & $2\times$ & $4\times$ & $8\times$ & $1\times$ & $2\times$ & $4\times$ & $8\times$ & $1\times$ & $2\times$ & $4\times$ & $8\times$ \\ \hline
    mip-NeRF 360~\cite{barron2022mipnerf360, multinerf2022} & 0.322 & 0.177 & 0.089 & 0.066 & 0.367 & 0.215 & 0.114 & 0.071 & 0.194 & 0.079 & 0.045 & 0.029 & 0.279 & 0.171 & 0.114 & 0.107 & 0.362 & 0.236 & 0.144 & 0.096 \\
mip-NeRF 360 + iNGP                                     & 0.313 & 0.166 & 0.128 & 0.169 & 0.344 & 0.192 & 0.124 & 0.137 & 0.192 & 0.079 & 0.107 & 0.176 & 0.254 & 0.156 & 0.137 & 0.180 & 0.344 & 0.223 & 0.171 & 0.182 \\
\hline
Our Model                                               & \textbf{0.222} & \textbf{0.112} & \textbf{0.061} & \textbf{0.041} & \textbf{0.287} & \textbf{0.156} & \textbf{0.083} & \textbf{0.050} & \textbf{0.129} & \textbf{0.055} & \textbf{0.030} & \textbf{0.020} & \textbf{0.206} & \textbf{0.122} & \textbf{0.077} & \textbf{0.057} & \textbf{0.263} & \textbf{0.163} & \textbf{0.103} & \textbf{0.068}
  \!\!\! \\
    \hline
    \end{tabular}
    }
    \resizebox{\linewidth}{!}{
    \begin{tabular}{|l||@{\,\,\,}c@{\,\,\,}c@{\,\,\,}c@{\,\,\,}c@{\,\,\,}|@{\,\,\,}c@{\,\,\,}c@{\,\,\,}c@{\,\,\,}c@{\,\,\,}|@{\,\,\,}c@{\,\,\,}c@{\,\,\,}c@{\,\,\,}c@{\,\,\,}|@{\,\,\,}c@{\,\,\,}c@{\,\,\,}c@{\,\,\,}c@{\,\,\,}|}
    \multicolumn{17}{c}{} \\
    \multicolumn{17}{c}{} \\
    \multicolumn{17}{c}{Indoors} \\
    \hline
    \bf{PSNR} 
    & \multicolumn{4}{c@{\,\,\,}|@{\,\,\,}}{\textit{room}} 
    & \multicolumn{4}{c@{\,\,\,}|@{\,\,\,}}{\textit{counter}}
    & \multicolumn{4}{c@{\,\,\,}|@{\,\,\,}}{\textit{kitchen}}
    & \multicolumn{4}{c@{\,\,\,}|}{\textit{bonsai}}  \\
    & $1\times$ & $2\times$ & $4\times$ & $8\times$ & $1\times$ & $2\times$ & $4\times$ & $8\times$ & $1\times$ & $2\times$ & $4\times$ & $8\times$ & $1\times$ & $2\times$ & $4\times$ & $8\times$ \\
    \hline
    mip-NeRF 360~\cite{barron2022mipnerf360, multinerf2022} & 31.44 & 32.53 & 33.17 & 32.96 & \textbf{29.30} & \textbf{30.12} & \textbf{30.81} & 30.52 & 31.90 & 33.39 & 34.69 & 34.92 & 32.85 & 33.97 & 34.63 & 33.80 \\
mip-NeRF 360 + iNGP                                     & 30.93 & 31.83 & 31.66 & 29.52 & 24.30 & 24.66 & 24.81 & 24.06 & 30.13 & 31.25 & 29.85 & 26.14 & 30.20 & 30.90 & 30.39 & 27.49 \\
\hline
Our Model                                               & \textbf{32.20} & \textbf{33.33} & \textbf{34.12} & \textbf{34.26} & 29.17 & 29.93 & 30.70 & \textbf{31.11} & \textbf{32.33} & \textbf{33.76} & \textbf{35.20} & \textbf{35.71} & \textbf{34.08} & \textbf{35.25} & \textbf{36.18} & \textbf{36.32}
  \!\!\! \\
    \hline
    \multicolumn{10}{c}{} \\
    \hline
    \bf{SSIM} & \multicolumn{4}{c@{\,\,\,}|@{\,\,\,}}{\textit{room}} 
    & \multicolumn{4}{c@{\,\,\,}|@{\,\,\,}}{\textit{counter}}
    & \multicolumn{4}{c@{\,\,\,}|@{\,\,\,}}{\textit{kitchen}}
    & \multicolumn{4}{c@{\,\,\,}|}{\textit{bonsai}}  \\
    & $1\times$ & $2\times$ & $4\times$ & $8\times$ & $1\times$ & $2\times$ & $4\times$ & $8\times$ & $1\times$ & $2\times$ & $4\times$ & $8\times$ & $1\times$ & $2\times$ & $4\times$ & $8\times$ \\ \hline
    mip-NeRF 360~\cite{barron2022mipnerf360, multinerf2022} & 0.906 & 0.944 & 0.963 & 0.967 & 0.887 & 0.916 & 0.936 & 0.942 & 0.916 & 0.949 & 0.968 & 0.975 & 0.935 & 0.959 & 0.969 & 0.968 \\
mip-NeRF 360 + iNGP                                     & 0.904 & 0.941 & 0.950 & 0.932 & 0.816 & 0.837 & 0.843 & 0.819 & 0.903 & 0.938 & 0.904 & 0.773 & 0.920 & 0.941 & 0.937 & 0.874 \\
\hline
Our Model                                               & \textbf{0.921} & \textbf{0.955} & \textbf{0.971} & \textbf{0.977} & \textbf{0.899} & \textbf{0.926} & \textbf{0.944} & \textbf{0.955} & \textbf{0.926} & \textbf{0.956} & \textbf{0.975} & \textbf{0.982} & \textbf{0.947} & \textbf{0.968} & \textbf{0.978} & \textbf{0.980}
  \!\!\! \\
    \hline
    \multicolumn{10}{c}{} \\
    \hline
    \bf{LPIPS} & \multicolumn{4}{c@{\,\,\,}|@{\,\,\,}}{\textit{room}} 
    & \multicolumn{4}{c@{\,\,\,}|@{\,\,\,}}{\textit{counter}}
    & \multicolumn{4}{c@{\,\,\,}|@{\,\,\,}}{\textit{kitchen}}
    & \multicolumn{4}{c@{\,\,\,}|}{\textit{bonsai}}  \\
    & $1\times$ & $2\times$ & $4\times$ & $8\times$ & $1\times$ & $2\times$ & $4\times$ & $8\times$ & $1\times$ & $2\times$ & $4\times$ & $8\times$ & $1\times$ & $2\times$ & $4\times$ & $8\times$ \\ \hline
    mip-NeRF 360~\cite{barron2022mipnerf360, multinerf2022} & 0.227 & 0.101 & 0.052 & 0.042 & 0.216 & 0.114 & 0.068 & 0.059 & 0.134 & 0.063 & 0.033 & 0.023 & 0.185 & 0.065 & 0.033 & 0.033 \\
mip-NeRF 360 + iNGP                                     & 0.220 & 0.105 & 0.072 & 0.095 & 0.275 & 0.195 & 0.163 & 0.182 & 0.145 & 0.074 & 0.084 & 0.188 & 0.190 & 0.082 & 0.063 & 0.126 \\
\hline
Our Model                                               & \textbf{0.199} & \textbf{0.084} & \textbf{0.041} & \textbf{0.028} & \textbf{0.189} & \textbf{0.095} & \textbf{0.055} & \textbf{0.039} & \textbf{0.117} & \textbf{0.055} & \textbf{0.028} & \textbf{0.018} & \textbf{0.173} & \textbf{0.052} & \textbf{0.023} & \textbf{0.017}
  \!\!\! \\
    \hline
    \end{tabular}
    }
    \caption{
    Multiscale per-scene performance on the dataset of ``360'' outdoor and indoor scenes from mip-NeRF 360 \cite{barron2022mipnerf360}.
    }
    \label{tab:scene_360_results_multi}
\end{table*}

\begin{table*}[b!]
    \centering
    \begin{tabular}{|l||cccccccc|c|}
    \hline
    \bf{PSNR} & \multicolumn{8}{c|}{} & \multicolumn{1}{c|}{} \\
     & \textit{chair} & \textit{drums} & \textit{ficus} & \textit{hotdog} & \textit{lego} & \textit{materials} & \textit{mic} & \textit{ship} & \textit{avg} \\\hline
NeRF~\cite{mildenhall2020,jaxnerf2020github}                        &                   34.17 &                   25.08 &                   30.39 &                   36.82 &                   33.31 &                   30.03 &                   34.78 &                   29.30 &                   31.74 \\
mip-NeRF~\cite{barron2021mipnerf}                                   & \cellcolor{orange}35.14 &                   25.48 & \cellcolor{orange}33.29 & \cellcolor{orange}37.48 & \cellcolor{orange}35.70 & \cellcolor{orange}30.71 & \cellcolor{orange}36.51 & \cellcolor{yellow}30.41 & \cellcolor{yellow}33.09 \\
mip-NeRF 360~\cite{barron2022mipnerf360, multinerf2022}, 256 hidden & \cellcolor{yellow}35.03 & \cellcolor{orange}25.73 &                   32.61 & \cellcolor{yellow}37.44 &    \cellcolor{red}36.10 & \cellcolor{yellow}30.31 & \cellcolor{yellow}36.22 &                   29.98 &                   32.93 \\
mip-NeRF 360~\cite{barron2022mipnerf360, multinerf2022}, 512 hidden &    \cellcolor{red}35.65 & \cellcolor{yellow}25.60 & \cellcolor{yellow}33.19 &    \cellcolor{red}37.71 &    \cellcolor{red}36.10 &                   29.90 &    \cellcolor{red}36.52 & \cellcolor{orange}31.26 &    \cellcolor{red}33.24 \\
\hline
Our Model                                                           &                   34.84 &    \cellcolor{red}25.84 &    \cellcolor{red}33.90 &                   37.14 & \cellcolor{yellow}34.84 &    \cellcolor{red}31.66 &                   35.15 &    \cellcolor{red}31.38 & \cellcolor{orange}33.10 \!\!\! \\
    \hline
    \multicolumn{10}{c}{} \\
    \hline
    \bf{SSIM} & \multicolumn{8}{c|}{} & \multicolumn{1}{c|}{} \\
    
 & \textit{chair} & \textit{drums} & \textit{ficus} & \textit{hotdog} & \textit{lego} & \textit{materials} & \textit{mic} & \textit{ship} & \textit{avg} \\\hline
NeRF~\cite{mildenhall2020,jaxnerf2020github}                        &                   0.975 &                   0.925 &                   0.967 &                   0.979 & \cellcolor{yellow}0.968 & \cellcolor{yellow}0.953 & \cellcolor{yellow}0.987 &                   0.869 &                   0.953 \\
mip-NeRF~\cite{barron2021mipnerf}                                   & \cellcolor{orange}0.981 & \cellcolor{yellow}0.932 & \cellcolor{orange}0.980 & \cellcolor{orange}0.982 & \cellcolor{orange}0.978 & \cellcolor{orange}0.959 &    \cellcolor{red}0.991 &                   0.882 & \cellcolor{orange}0.961 \\
mip-NeRF 360~\cite{barron2022mipnerf360, multinerf2022}, 256 hidden & \cellcolor{yellow}0.980 & \cellcolor{orange}0.934 &                   0.977 & \cellcolor{yellow}0.981 &    \cellcolor{red}0.980 & \cellcolor{yellow}0.953 & \cellcolor{orange}0.990 & \cellcolor{yellow}0.883 & \cellcolor{yellow}0.960 \\
mip-NeRF 360~\cite{barron2022mipnerf360, multinerf2022}, 512 hidden &    \cellcolor{red}0.983 &                   0.931 & \cellcolor{yellow}0.979 & \cellcolor{orange}0.982 &    \cellcolor{red}0.980 &                   0.949 &    \cellcolor{red}0.991 & \cellcolor{orange}0.893 & \cellcolor{orange}0.961 \\
\hline
Our Model                                                           &    \cellcolor{red}0.983 &    \cellcolor{red}0.944 &    \cellcolor{red}0.985 &    \cellcolor{red}0.984 &    \cellcolor{red}0.980 &    \cellcolor{red}0.969 &    \cellcolor{red}0.991 &    \cellcolor{red}0.929 &    \cellcolor{red}0.971  \!\!\! \\
    \hline
    \multicolumn{10}{c}{} \\
    \hline
    \bf{LPIPS} & \multicolumn{8}{c|}{} & \multicolumn{1}{c|}{} \\
     & \textit{chair} & \textit{drums} & \textit{ficus} & \textit{hotdog} & \textit{lego} & \textit{materials} & \textit{mic} & \textit{ship} & \textit{avg} \\\hline
NeRF~\cite{mildenhall2020,jaxnerf2020github}                        &                   0.026 &                   0.071 &                   0.032 &                   0.030 &                   0.031 & \cellcolor{yellow}0.047 &                   0.012 &                   0.150 &                   0.050 \\
mip-NeRF~\cite{barron2021mipnerf}                                   & \cellcolor{yellow}0.021 & \cellcolor{yellow}0.065 & \cellcolor{orange}0.020 & \cellcolor{yellow}0.027 & \cellcolor{yellow}0.021 & \cellcolor{orange}0.040 & \cellcolor{orange}0.009 &                   0.138 & \cellcolor{yellow}0.043 \\
mip-NeRF 360~\cite{barron2022mipnerf360, multinerf2022}, 256 hidden & \cellcolor{yellow}0.021 & \cellcolor{orange}0.064 &                   0.024 & \cellcolor{yellow}0.027 &    \cellcolor{red}0.018 & \cellcolor{yellow}0.047 & \cellcolor{yellow}0.011 & \cellcolor{yellow}0.135 & \cellcolor{yellow}0.043 \\
mip-NeRF 360~\cite{barron2022mipnerf360, multinerf2022}, 512 hidden & \cellcolor{orange}0.018 &                   0.069 & \cellcolor{yellow}0.022 & \cellcolor{orange}0.024 &    \cellcolor{red}0.018 &                   0.053 & \cellcolor{yellow}0.011 & \cellcolor{orange}0.119 & \cellcolor{orange}0.042 \\
\hline
Our Model                                                           &    \cellcolor{red}0.017 &    \cellcolor{red}0.050 &    \cellcolor{red}0.015 &    \cellcolor{red}0.020 & \cellcolor{orange}0.019 &    \cellcolor{red}0.032 &    \cellcolor{red}0.007 &    \cellcolor{red}0.091 &    \cellcolor{red}0.031 \!\!\! \\
    \hline
    \end{tabular}
    \caption{
    Per-scene and average performance on the Blender dataset from NeRF~\cite{mildenhall2020}.
    }
    \label{tab:blender_results}
\end{table*}

\end{document}